\documentclass[11pt,a4paper]{article}
\pdfoutput=1
\usepackage[hyperref]{acl2018}
\usepackage[utf8]{inputenc}
\usepackage[T1]{fontenc}
\usepackage{times}
\usepackage{latexsym}
\usepackage[tone]{tipa}
\usepackage{rotating}
\usepackage[small]{caption}
\usepackage{xspace}
\usepackage{subcaption}

\usepackage{algorithm}
\usepackage[noend]{algpseudocode}

\usepackage{url}
\usepackage{microtype}

\usepackage{adjustbox}
\usepackage{booktabs}

\usepackage{amsmath}
\usepackage{amssymb}
\usepackage{amsfonts}
\usepackage{amsthm}
\usepackage{mathtools}
\usepackage{tikz}
\usepackage{xfrac}

\usetikzlibrary{bayesnet}

\usepackage[outline]{contour}
\usepackage{varwidth}

\newcommand{\rbox}[1]{{\footnotesize \begin{rotate}{90}#1\end{rotate}}}

\newcommand{\PL}{{\setlength{\fboxsep}{0pt}\fbox{\includegraphics[height=0.30cm,width=0.45cm]{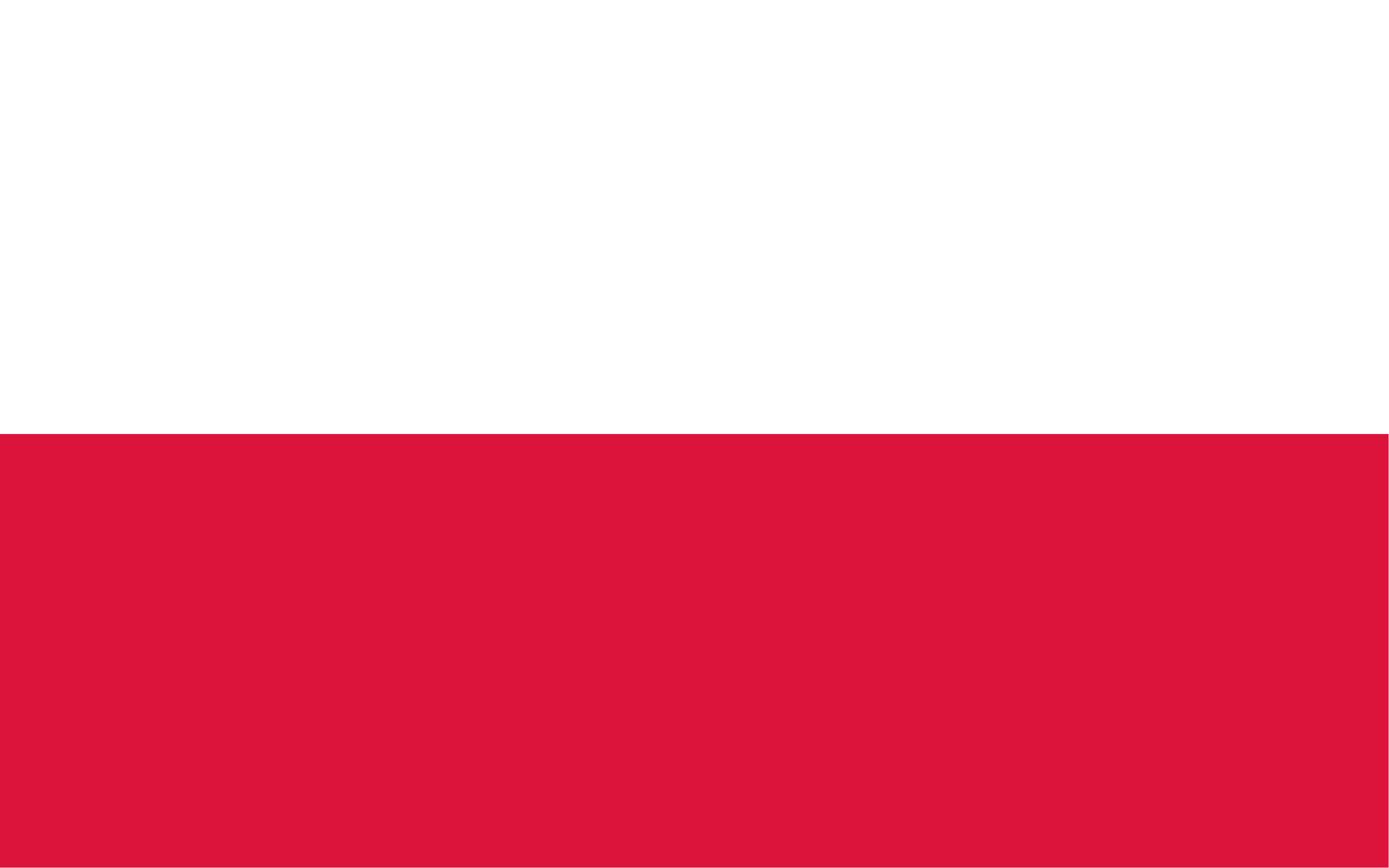}}}}
\newcommand{\BG}{{\setlength{\fboxsep}{0pt}\fbox{\includegraphics[height=0.30cm,width=0.45cm]{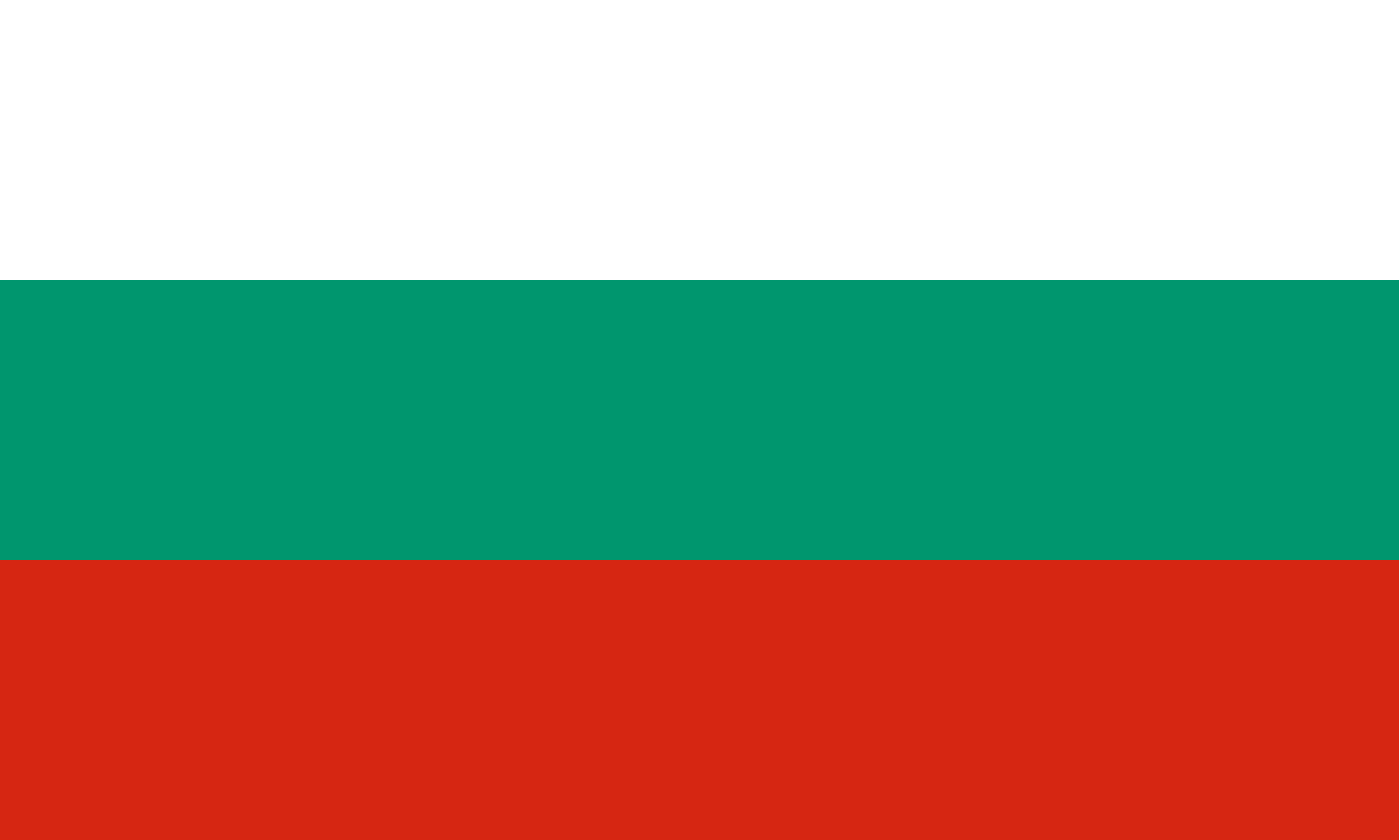}}}}
\newcommand{\SI}{{\setlength{\fboxsep}{0pt}\fbox{\includegraphics[height=0.30cm,width=0.45cm]{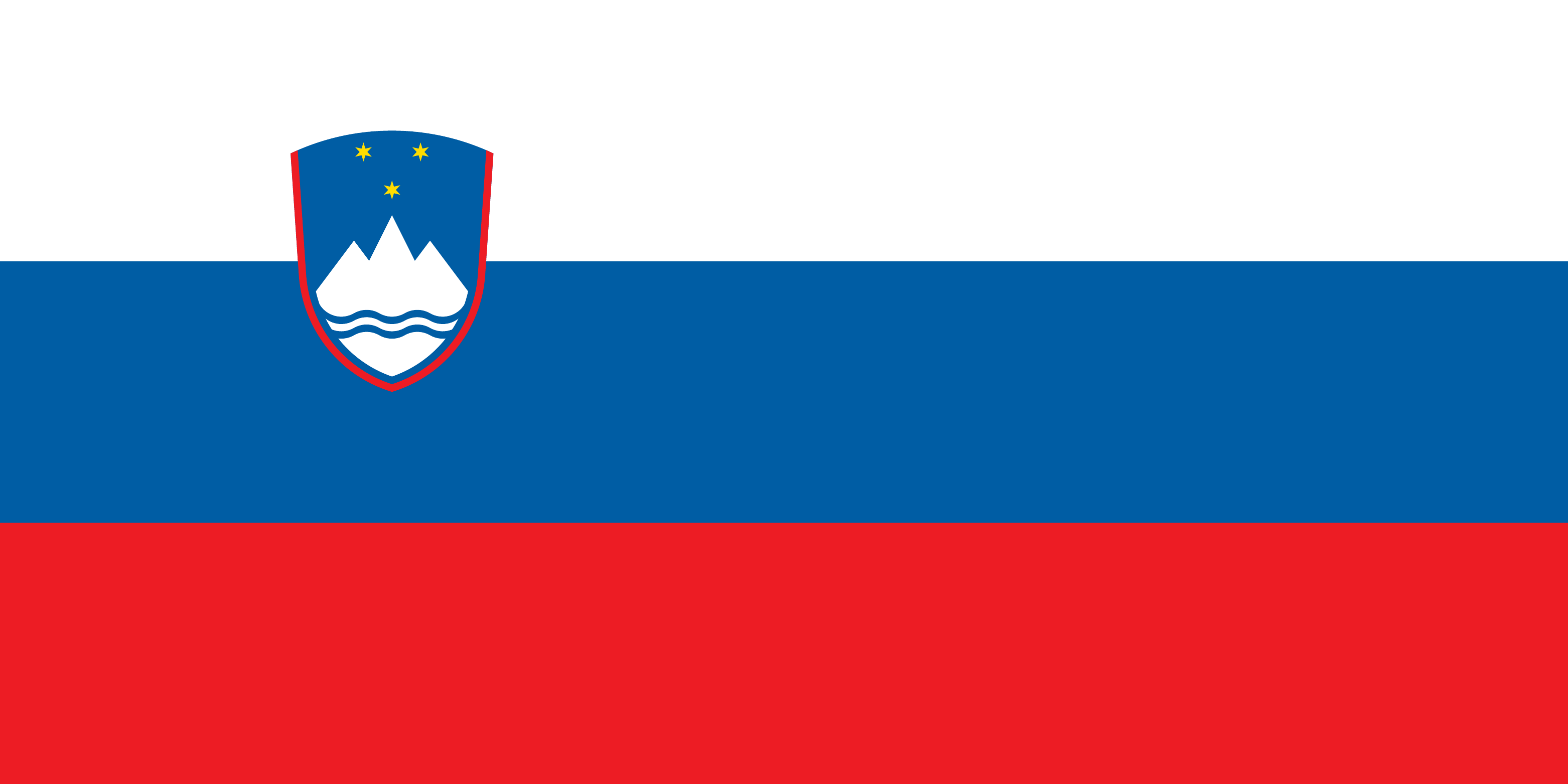}}}}
\newcommand{\CA}{{\setlength{\fboxsep}{0pt}\fbox{\includegraphics[height=0.30cm,width=0.45cm]{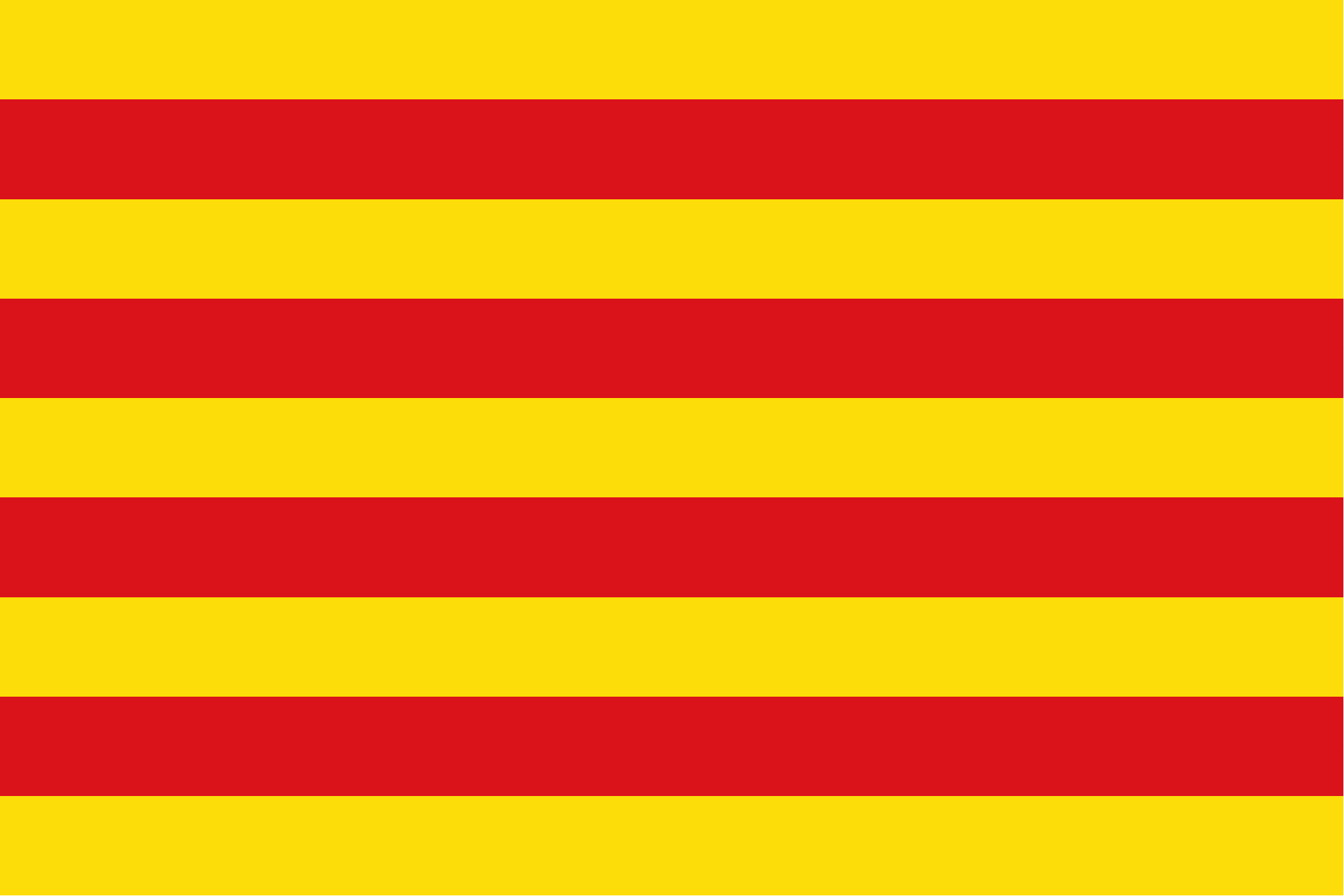}}}}
\newcommand{\LA}{{\setlength{\fboxsep}{0pt}\fbox{\includegraphics[height=0.30cm,width=0.45cm]{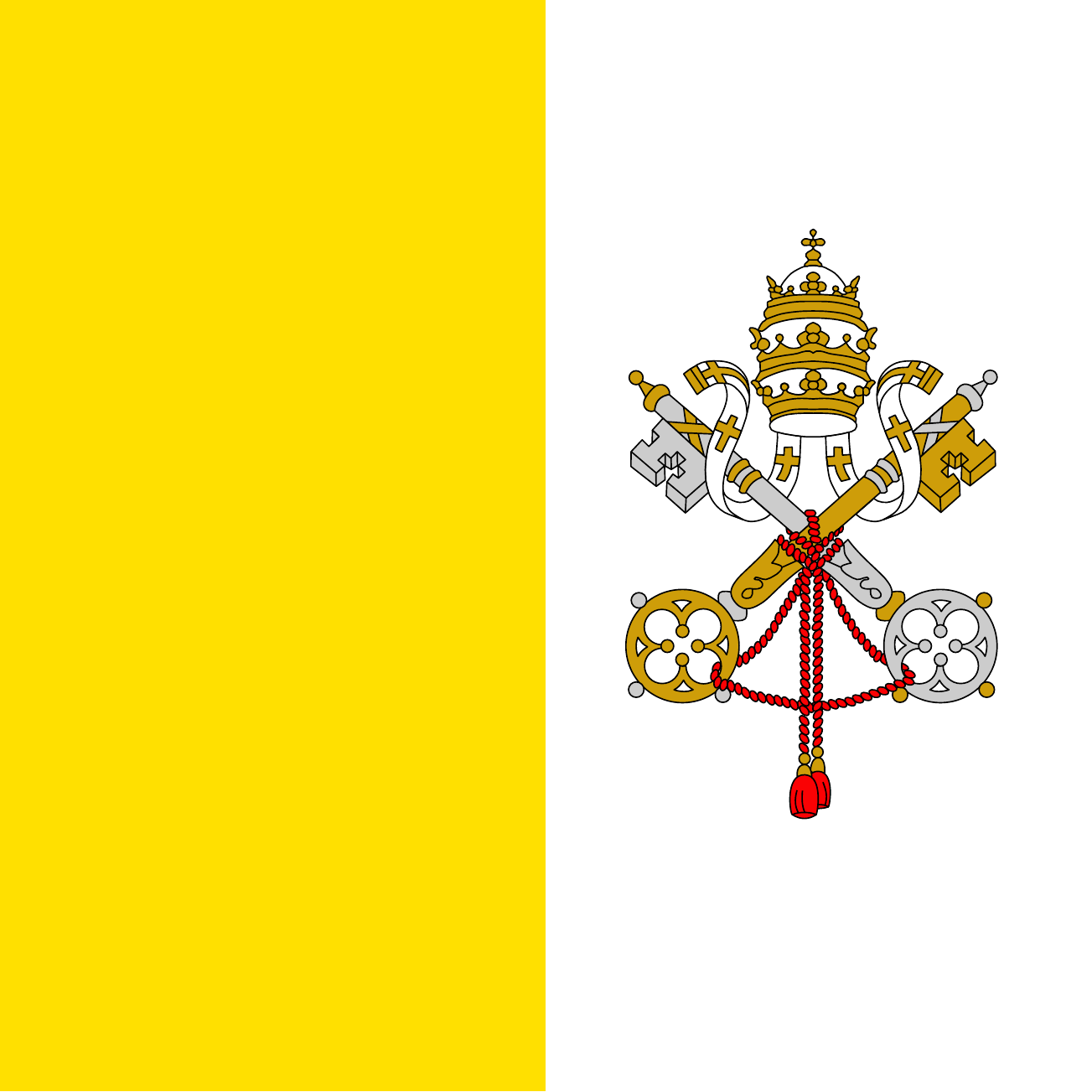}}}}
\newcommand{\ES}{{\setlength{\fboxsep}{0pt}\fbox{\includegraphics[height=0.30cm,width=0.45cm]{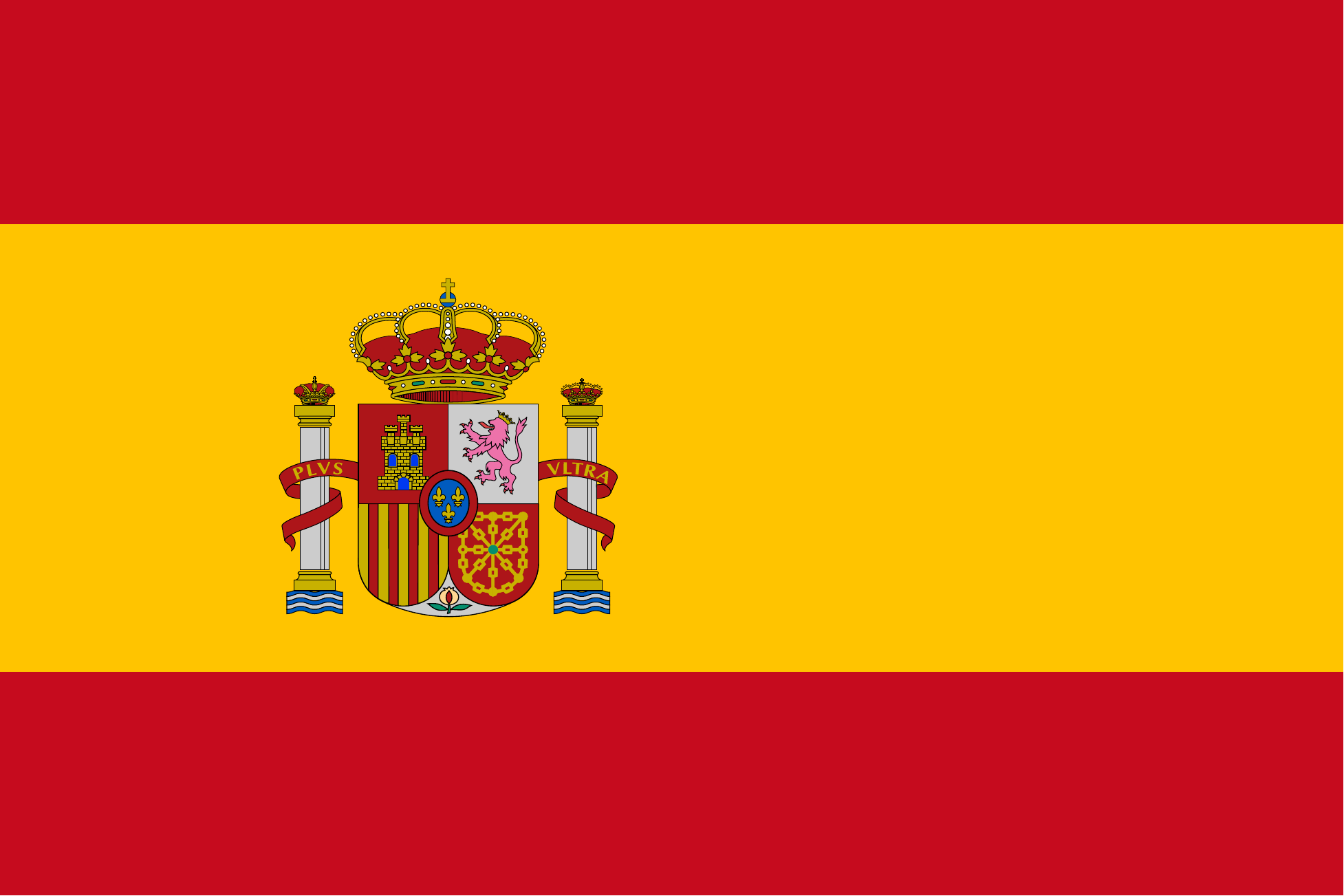}}}}
\newcommand{\PT}{{\setlength{\fboxsep}{0pt}\fbox{\includegraphics[height=0.30cm,width=0.45cm]{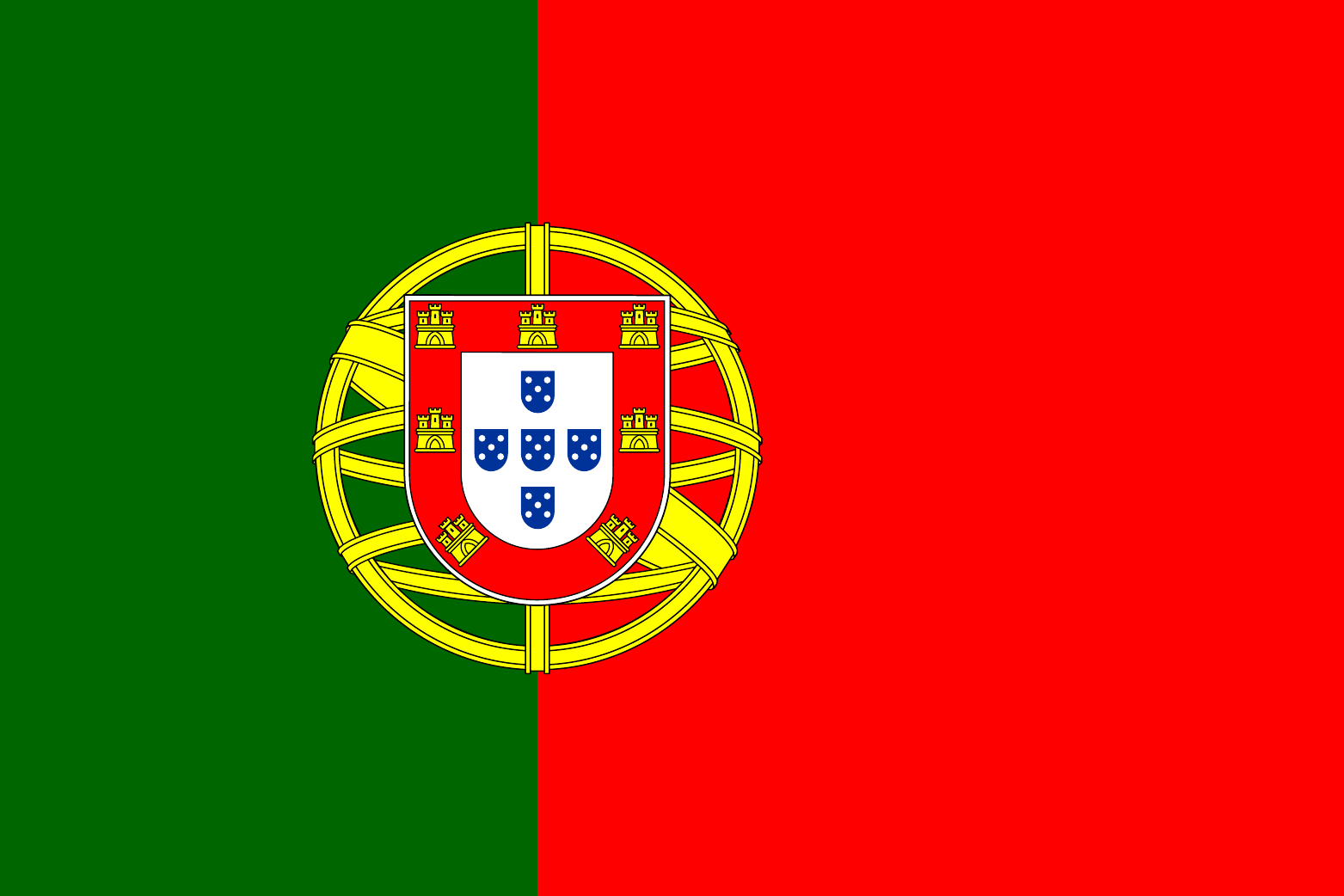}}}}
\newcommand{\IT}{{\setlength{\fboxsep}{0pt}\fbox{\includegraphics[height=0.30cm,width=0.45cm]{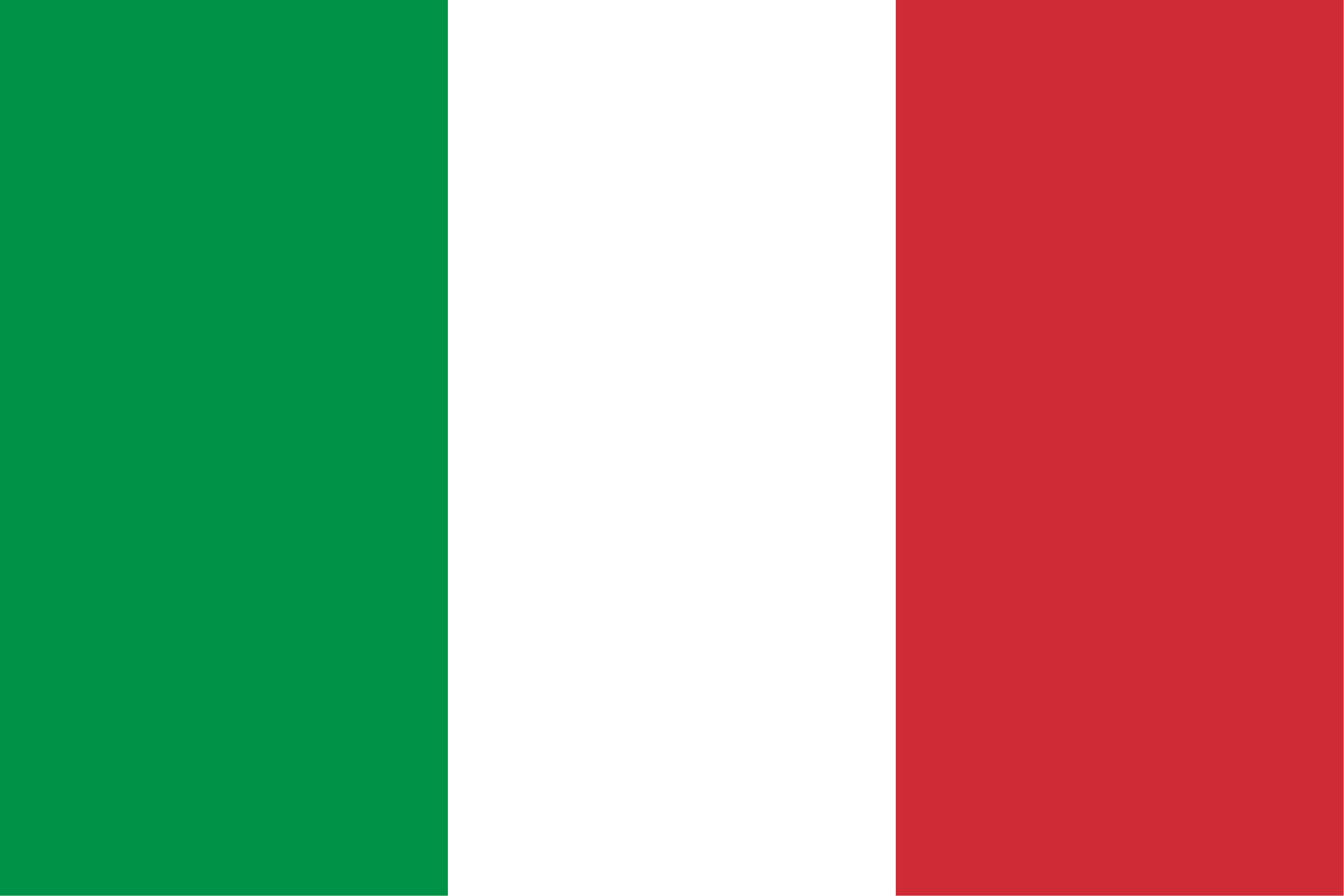}}}}
\newcommand{\FR}{{\setlength{\fboxsep}{0pt}\fbox{\includegraphics[height=0.30cm,width=0.45cm]{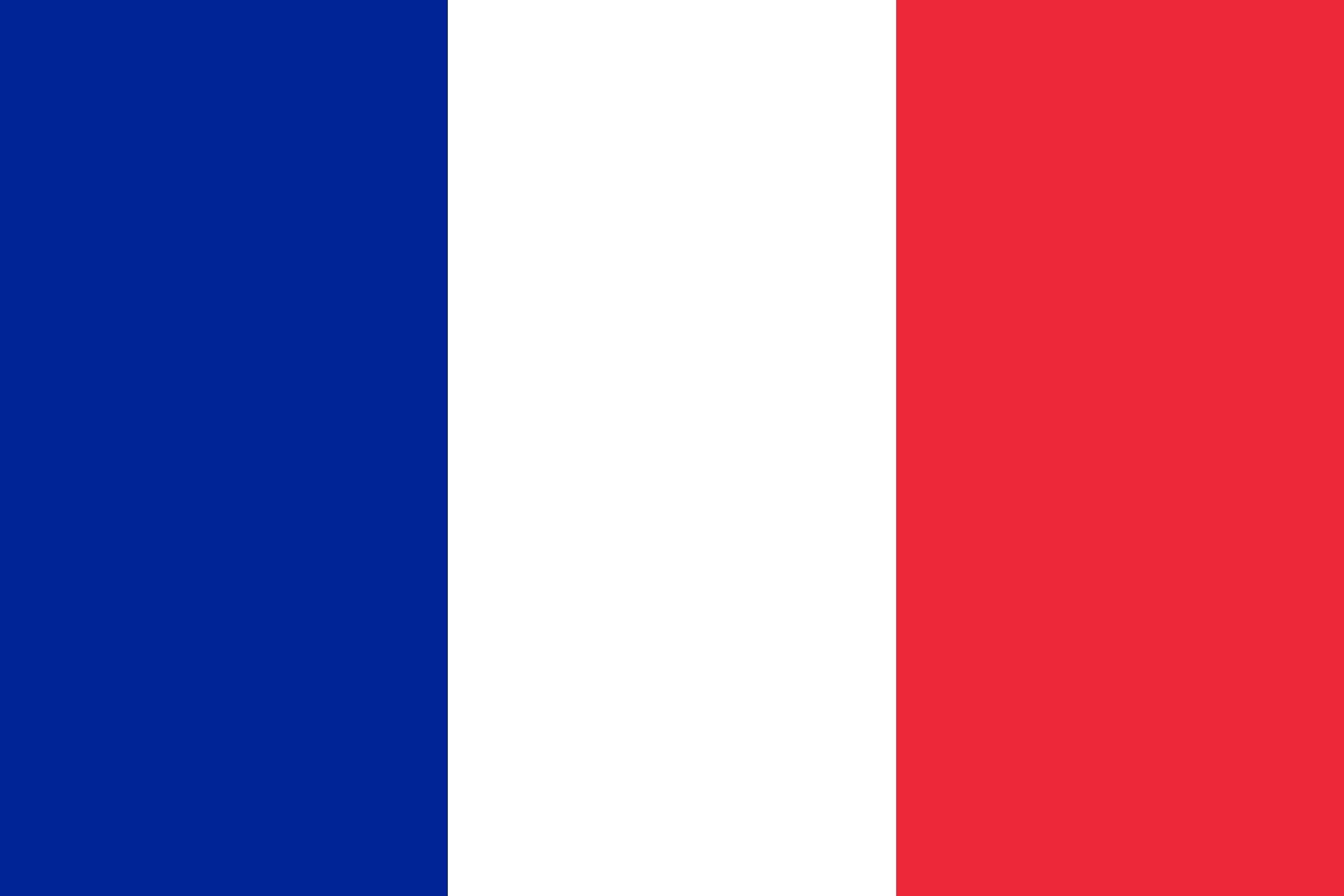}}}}
\newcommand{\RO}{{\setlength{\fboxsep}{0pt}\fbox{\includegraphics[height=0.30cm,width=0.45cm]{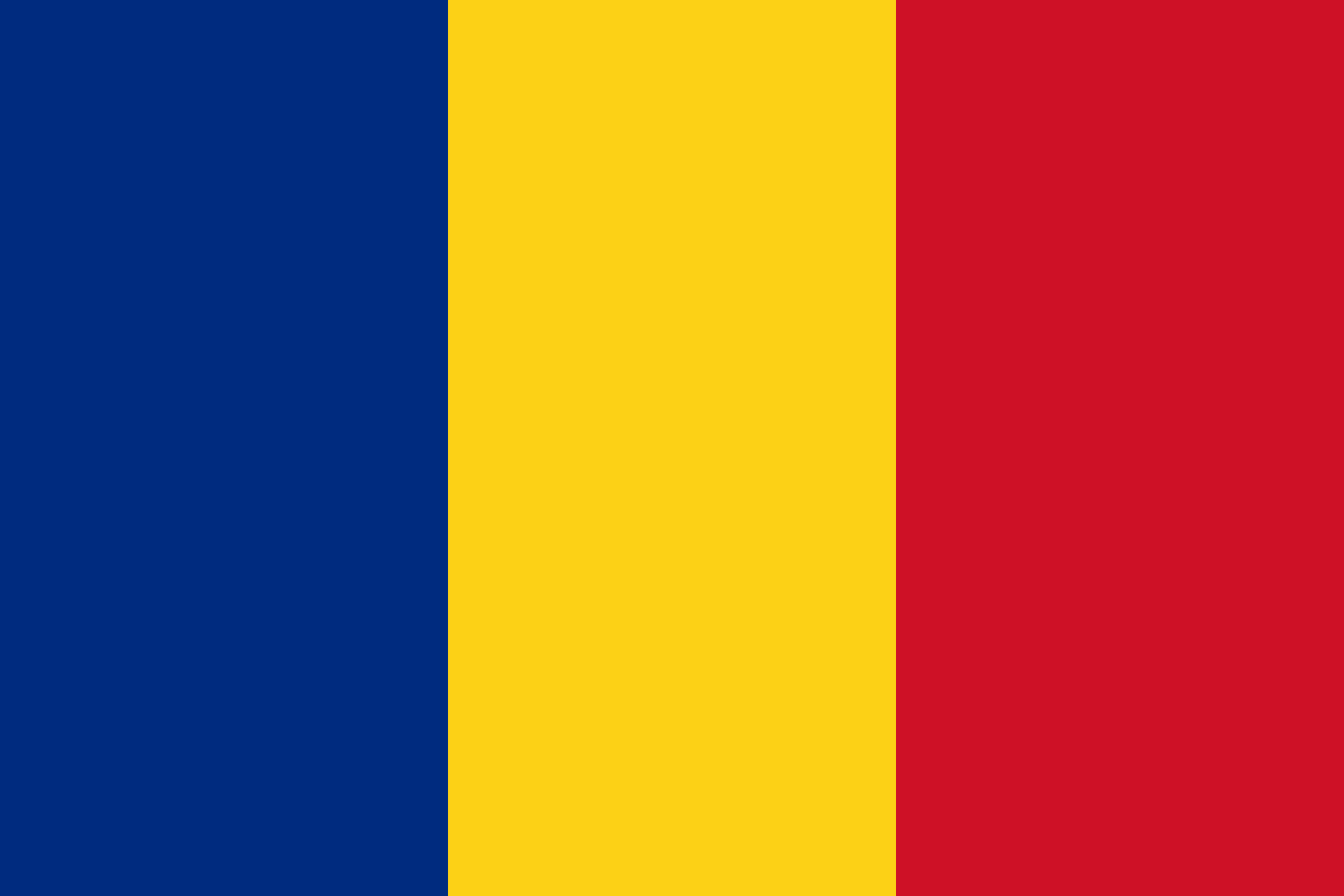}}}}
\newcommand{\SV}{{\setlength{\fboxsep}{0pt}\fbox{\includegraphics[height=0.30cm,width=0.45cm]{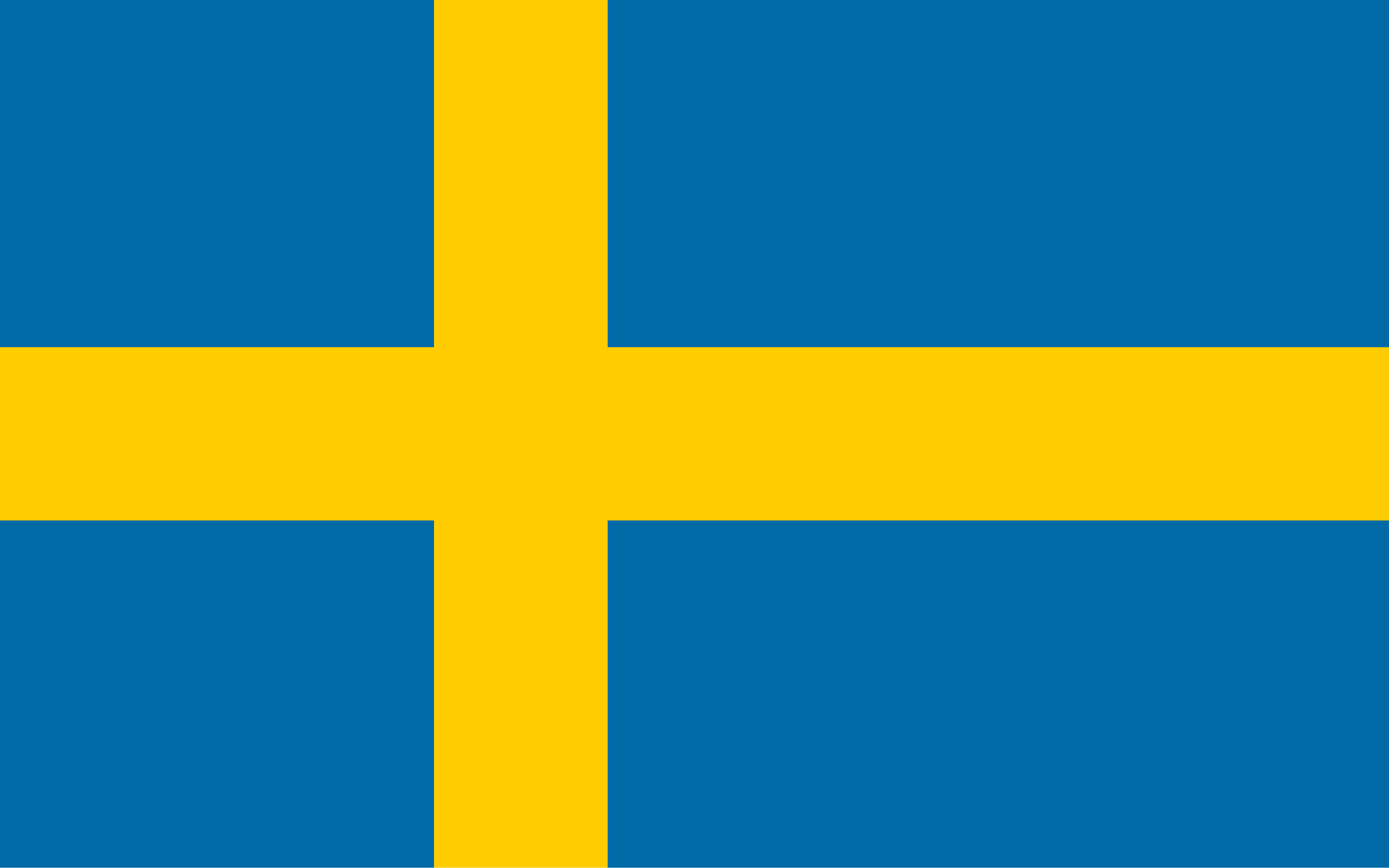}}}}
\newcommand{\NO}{{\setlength{\fboxsep}{0pt}\fbox{\includegraphics[height=0.30cm,width=0.45cm]{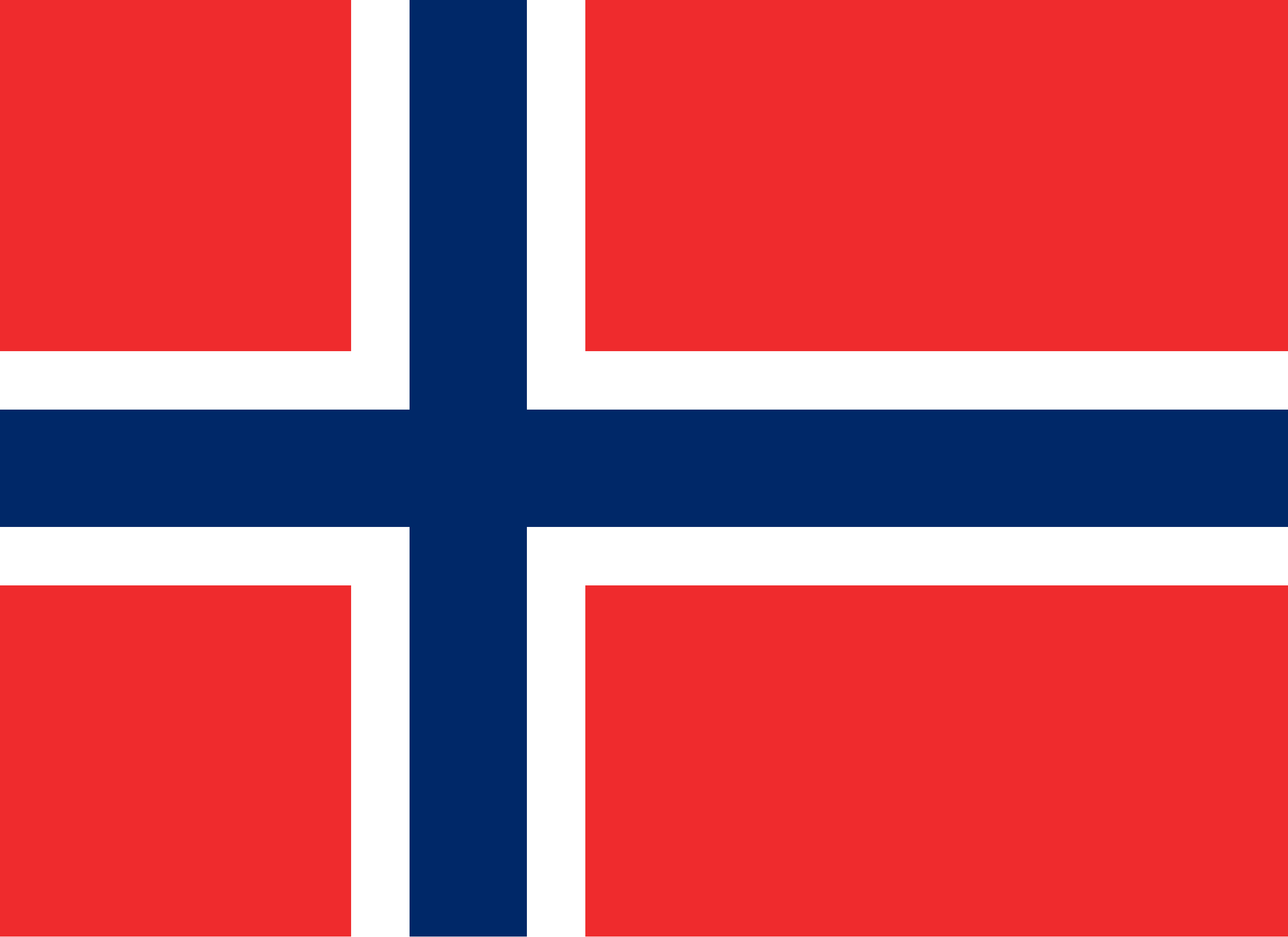}}}}
\newcommand{\DA}{{\setlength{\fboxsep}{0pt}\fbox{\includegraphics[height=0.30cm,width=0.45cm]{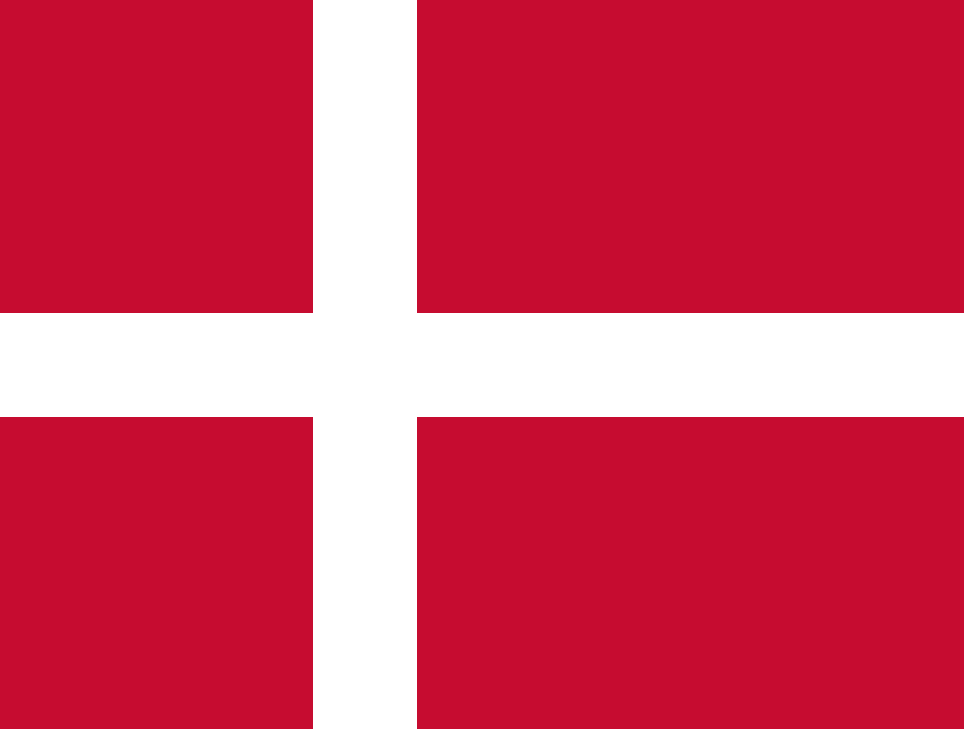}}}}
\newcommand{\NL}{{\setlength{\fboxsep}{0pt}\fbox{\includegraphics[height=0.30cm,width=0.45cm]{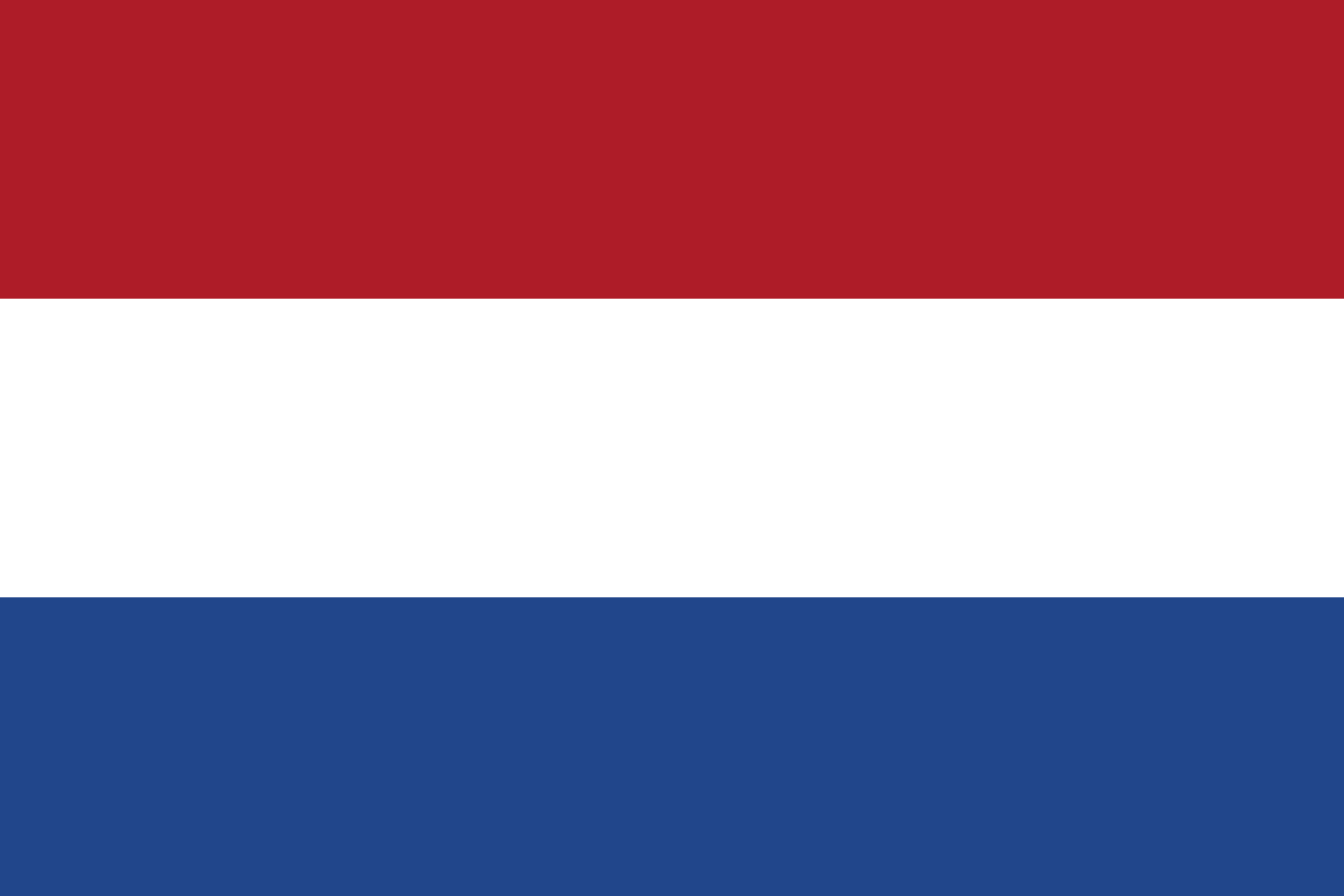}}}}
\newcommand{\HU}{{\setlength{\fboxsep}{0pt}\fbox{\includegraphics[height=0.30cm,width=0.45cm]{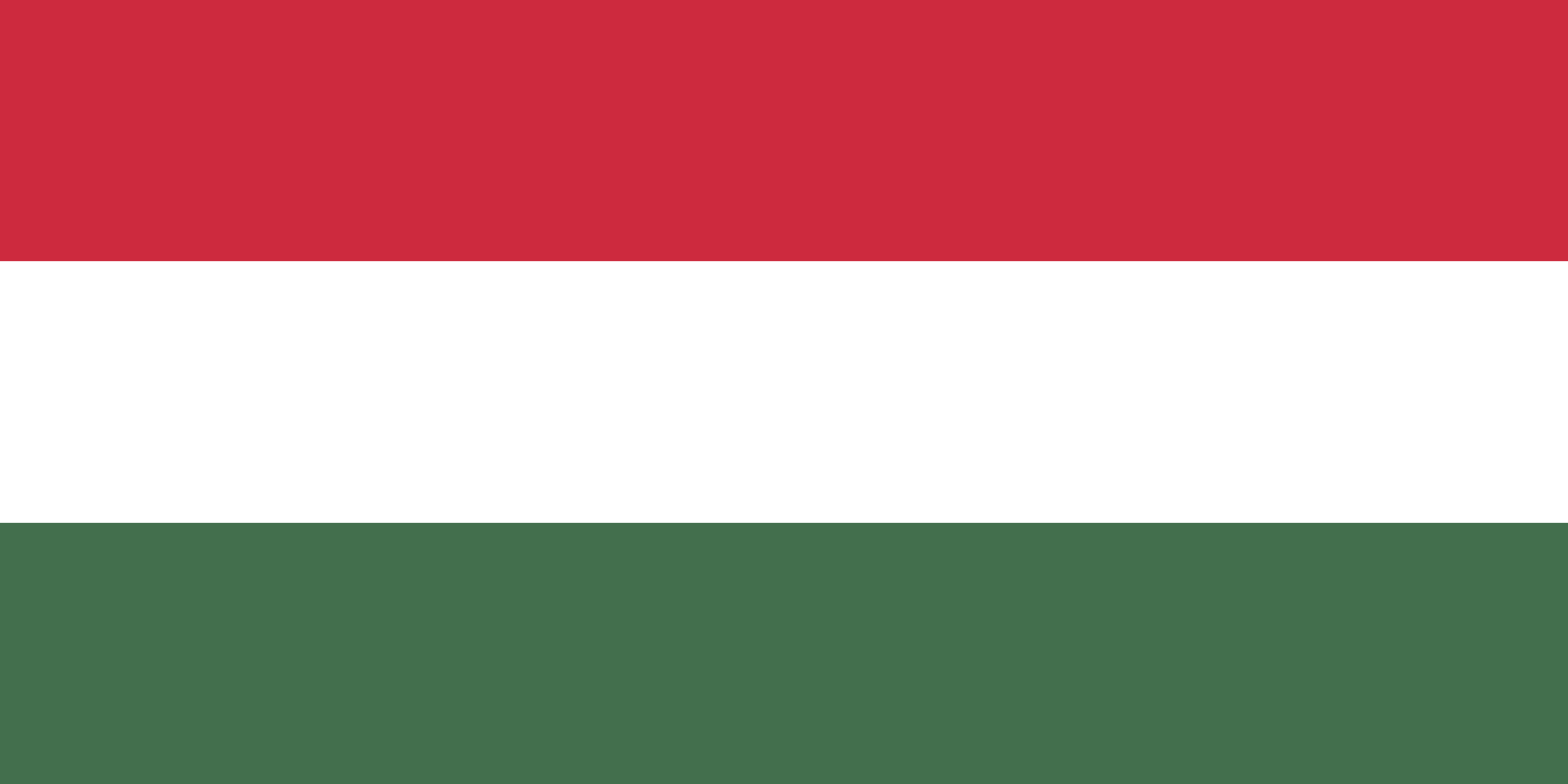}}}}
\newcommand{\FI}{{\setlength{\fboxsep}{0pt}\fbox{\includegraphics[height=0.30cm,width=0.45cm]{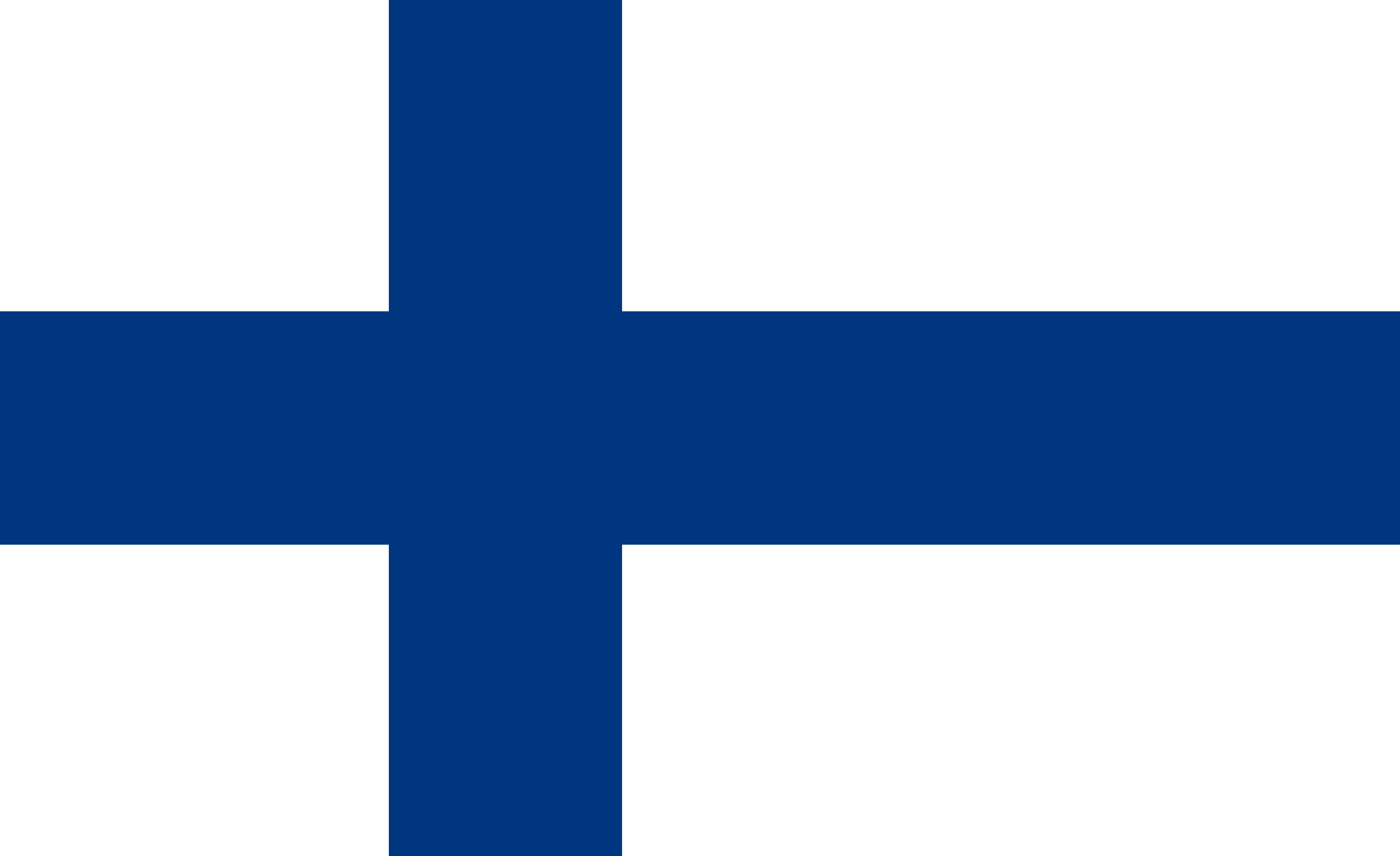}}}}
\newcommand{\ET}{{\setlength{\fboxsep}{0pt}\fbox{\includegraphics[height=0.30cm,width=0.45cm]{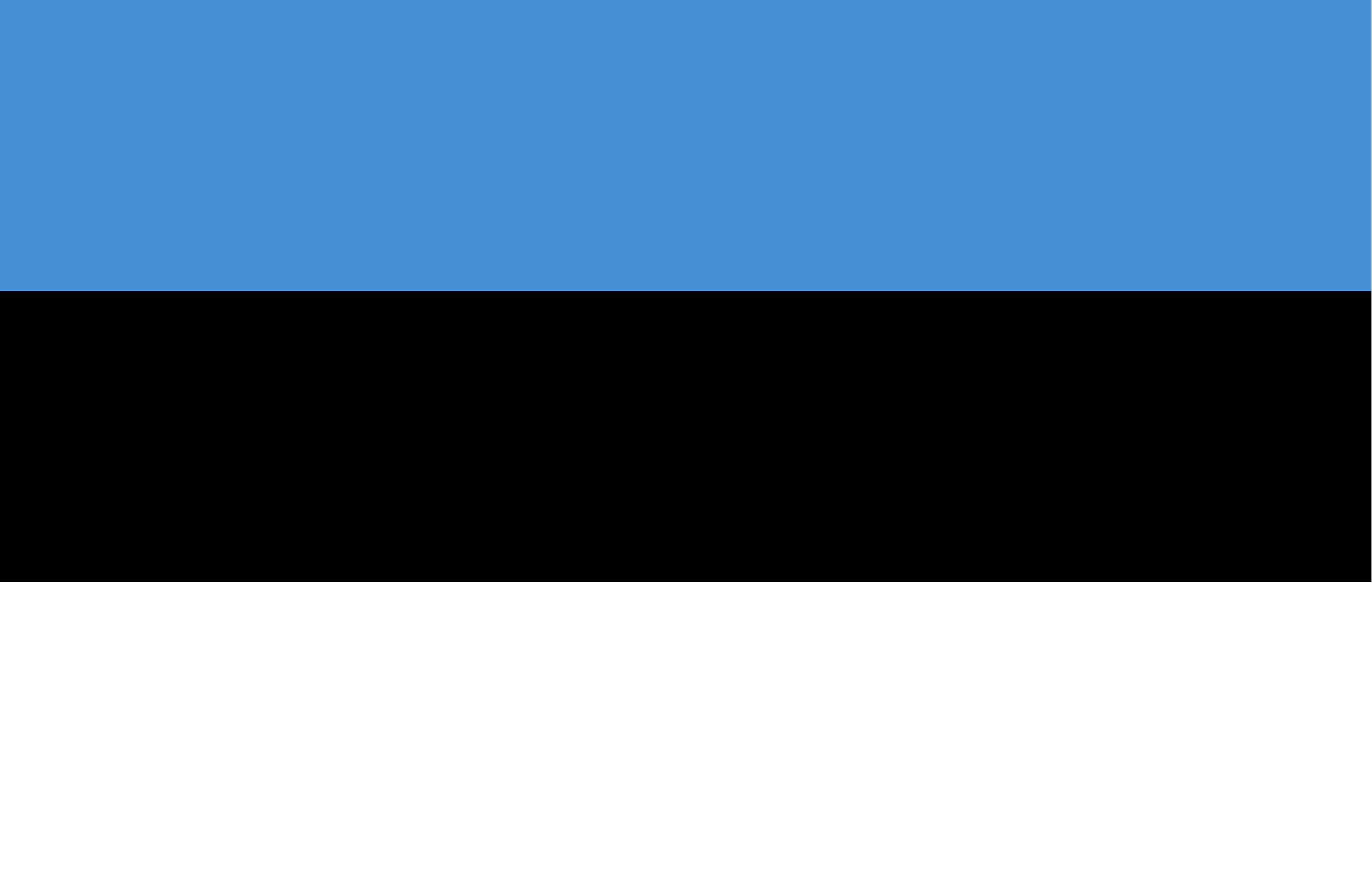}}}}
\newcommand{\EU}{{\setlength{\fboxsep}{0pt}\fbox{\includegraphics[height=0.30cm,width=0.45cm]{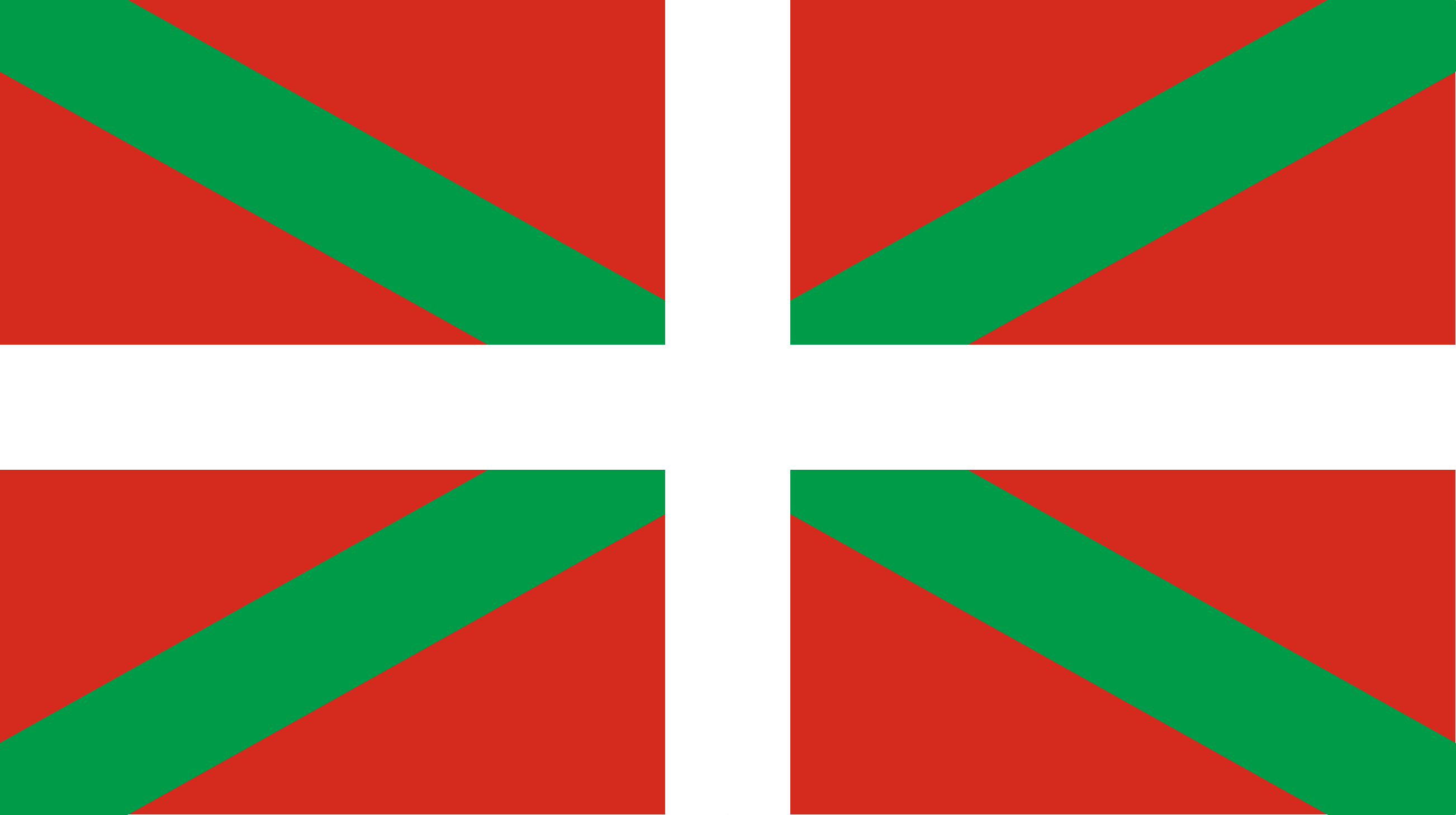}}}}
\newcommand{\LV}{{\setlength{\fboxsep}{0pt}\fbox{\includegraphics[height=0.30cm,width=0.45cm]{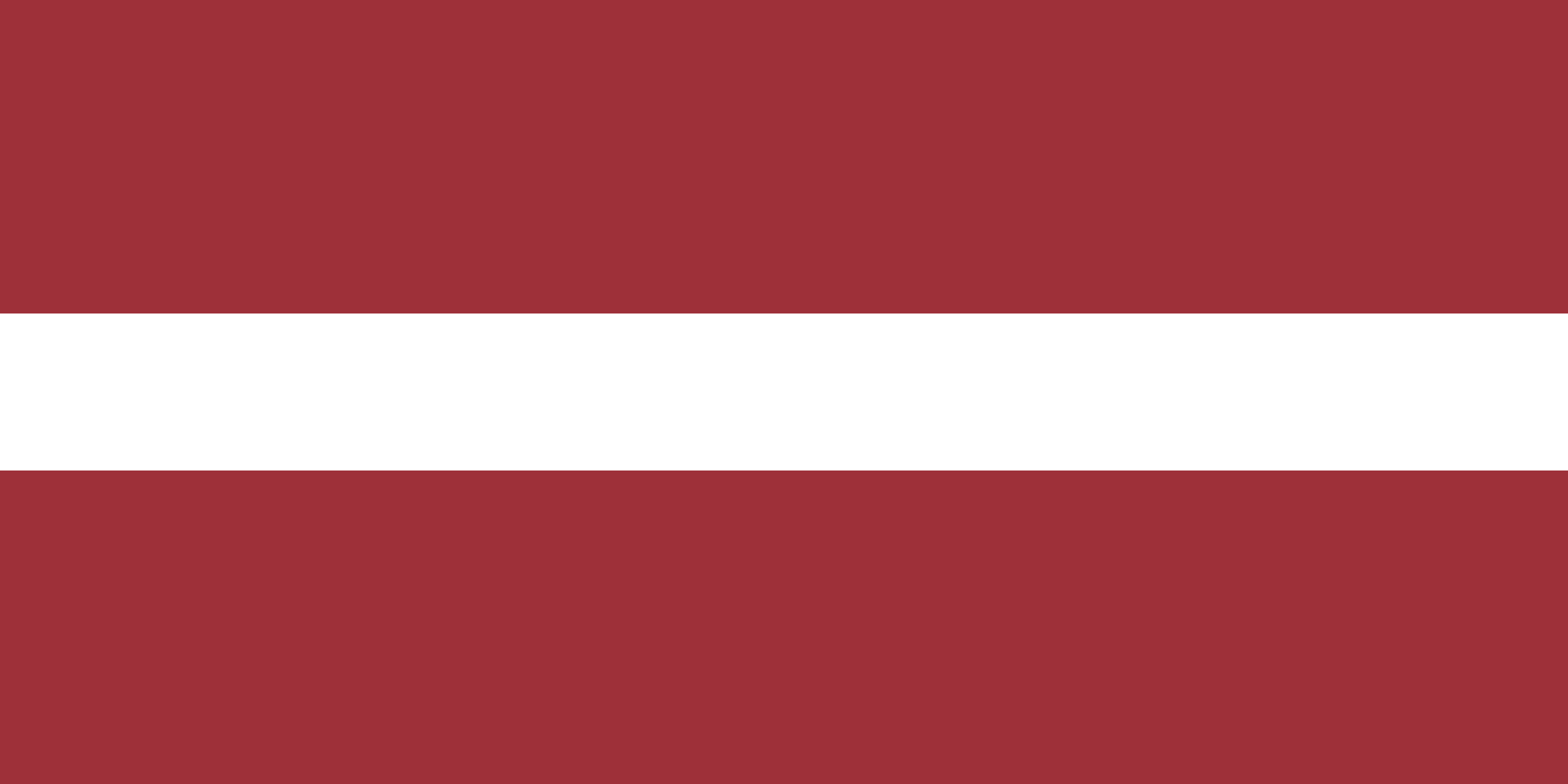}}}}
\newcommand{\TR}{{\setlength{\fboxsep}{0pt}\fbox{\includegraphics[height=0.30cm,width=0.45cm]{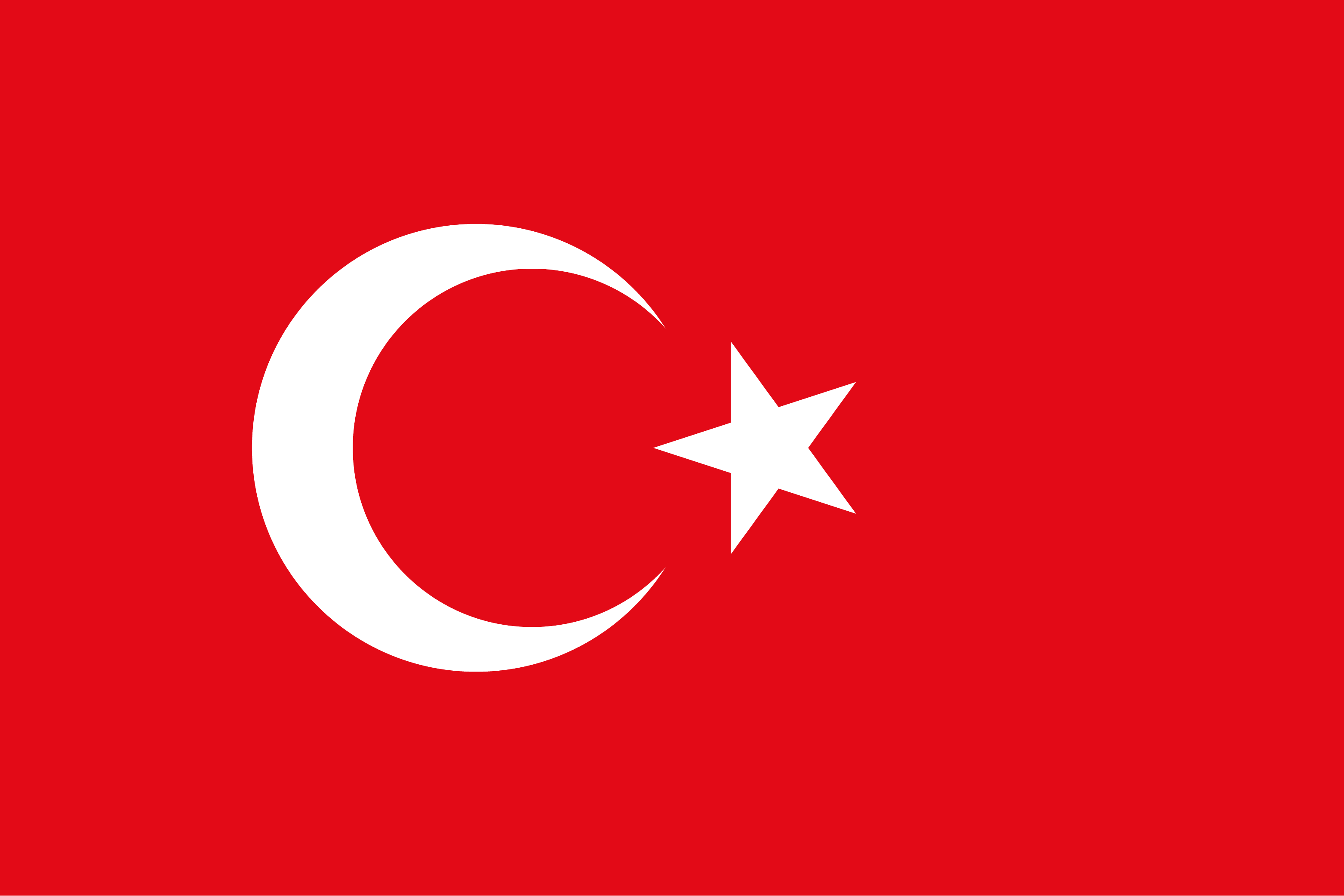}}}}
\newcommand{\HE}{{\setlength{\fboxsep}{0pt}\fbox{\includegraphics[height=0.30cm,width=0.45cm]{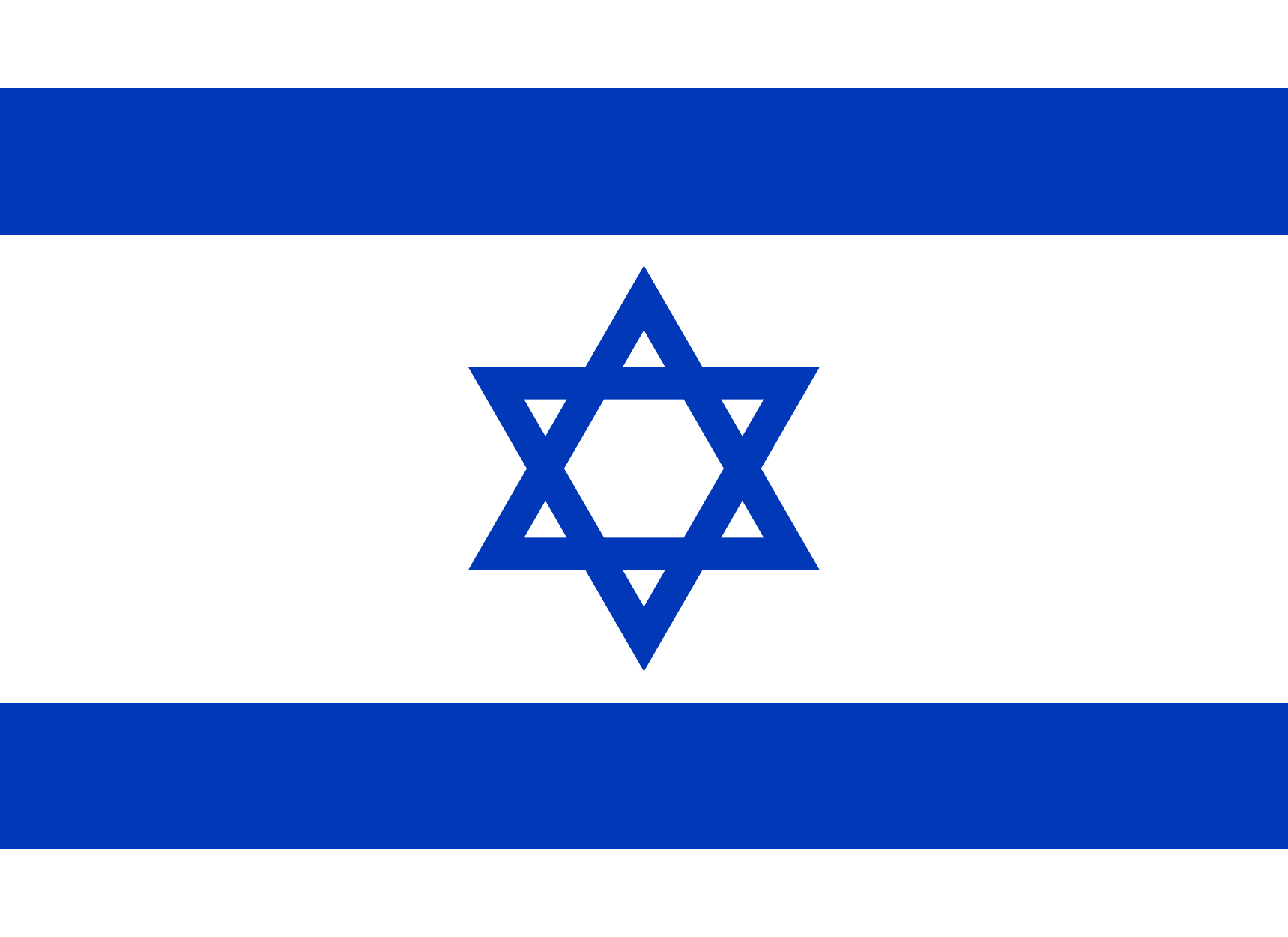}}}}
\newcommand{\AR}{{\setlength{\fboxsep}{0pt}\fbox{\includegraphics[height=0.30cm,width=0.45cm]{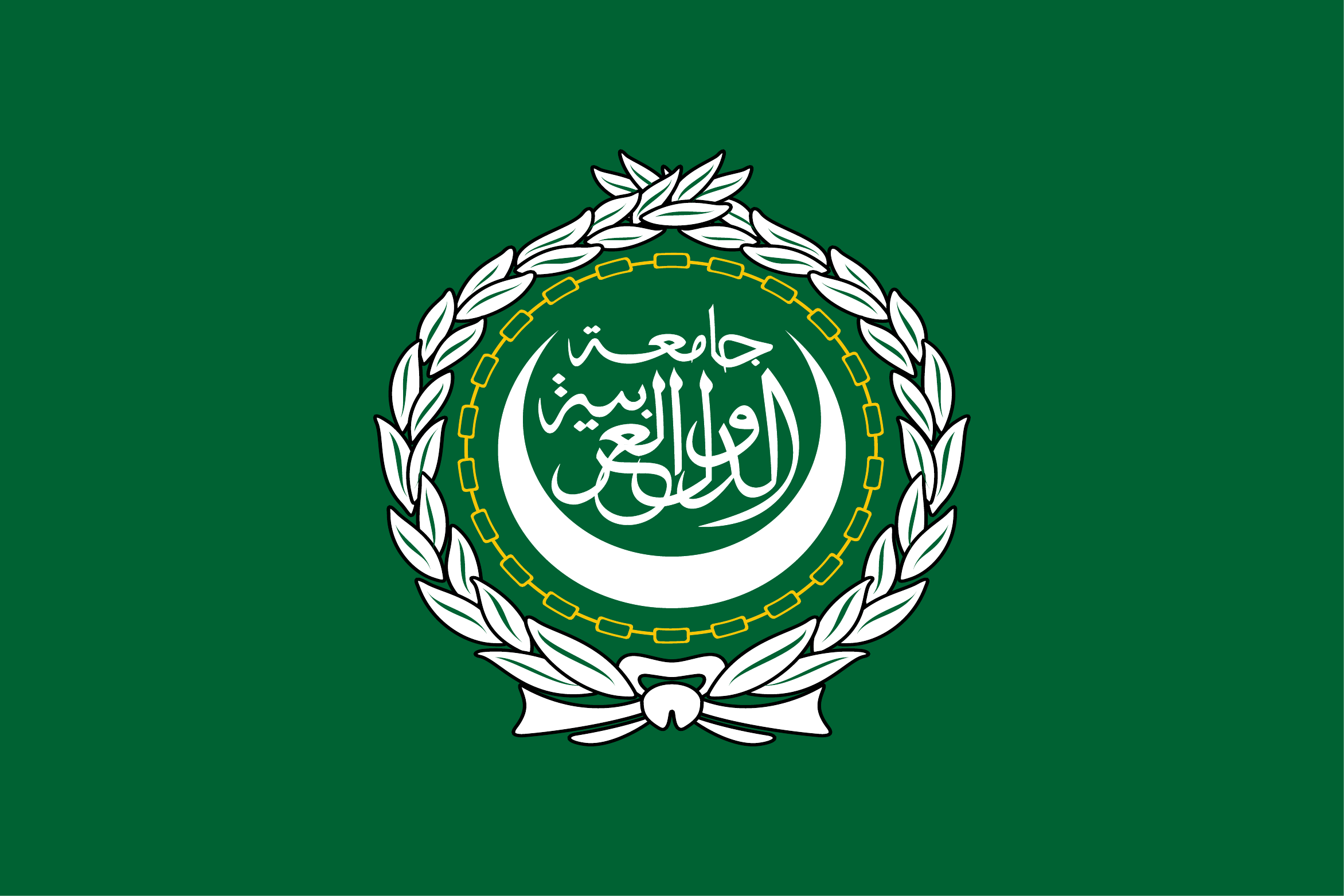}}}}

\newcommand*{\numberingBlue}[1]{%
  \protect\tikz[baseline={([yshift=-1.5pt]n.base)}]%
  \protect\node[fill=cbcol3,shape=circle,inner sep=1pt,draw](n){\tiny #1};}

\newcommand*{\numberingRed}[1]{%
  \protect\tikz[baseline={([yshift=-1.5pt]n.base)}]%
  \protect\node[fill=cbcol1 ,shape=circle,inner sep=1pt,draw](n){\tiny #1};}

\newcommand*{\numberingGreen}[1]{%
  \protect\tikz[baseline={([yshift=-1.5pt]n.base)}]%
  \protect\node[fill=cbcol2,shape=circle,inner sep=1pt,draw](n){\tiny #1};}

\newcommand*{\numberingBlueB}[1]{%
  \protect\tikz[baseline={([yshift=-1.5pt]n.base)}]%
  \protect\node[fill=cbcol3,shape=circle,inner sep=1pt,draw](n){\small #1};}

\newcommand*{\numberingRedB}[1]{%
  \protect\tikz[baseline={([yshift=-1.5pt]n.base)}]%
  \protect\node[fill=cbcol1 ,shape=circle,inner sep=1pt,draw](n){\small #1};}

\newcommand*{\numberingGreenB}[1]{%
  \protect\tikz[baseline={([yshift=-1.5pt]n.base)}]%
  \protect\node[fill=cbcol2,shape=circle,inner sep=1pt,draw](n){\small #1};}

\newcommand{\Dlabeled}{{\cal D}_\textit{labeled}}
\newcommand{\Dunlabeled}{{\cal D}_\textit{unlabeled}}

\usepackage{xcolor}

\definecolor{cbcol1}{rgb}{1.00,0.75,0.75}
\definecolor{cbcol2}{rgb}{0.75,1.00,0.75}
\definecolor{cbcol3}{rgb}{0.75,0.75,1.00}

\usepackage{courier}  
\newcommand{\defn}[1]{\textbf{#1}}
\newcommand{\word}[1]{\textsf{\small #1}}  
\newcommand{\pos}[1]{\textsc{#1}}  
\newcommand{\slot}[1]{\textsc{#1}} 
\newcommand{\attval}[2]{\mbox{\slot{#1}$=$\slot{#2}}}
\newcommand{\size}[1]{\ensuremath{\left|#1\right|}}

\newcommand{\angles}[1]{\langle #1 \rangle}
\newcommand{\brackets}[1]{\big[ #1 \big]}

\usepackage{cleveref}
\crefname{section}{\S}{\S\S}
\Crefname{section}{\S}{\S\S}
\crefname{table}{Tab.}{}
\crefname{figure}{Fig.}{}
\crefname{algorithm}{Alg.}{}
\crefname{equation}{Eq.}{}
\crefname{appendix}{App.}{}
\crefformat{section}{\S#2#1#3}  

\newcommand{\vm}{{\boldsymbol m}}
\newcommand{\vl}{{\boldsymbol \ell}}
\newcommand{\vtheta}{{\boldsymbol \theta}}

\newcommand{\vf}{{\boldsymbol f}}

\newcommand{\vell}{{\boldsymbol \ell}}
\newcommand{\vphi}{{\boldsymbol \phi}}

\newcommand{\vDelta}{{\boldsymbol \Delta}}

\newcommand{\tilvf}{\tilde{\vf}}
\newcommand{\tilvm}{\tilde{\vm}}
\newcommand{\tilvell}{\tilde{\vell}}

\newcommand{\qphi}{q_{\vphi}}
\newcommand{\ptheta}{p_{\vtheta}}

\newcommand{\Ddreamt}{\widetilde{\mathcal{D}}_\textit{sleep}}
\newcommand{\Dwake}{\widetilde{\mathcal{D}}_\textit{wake}}

\newcommand\lemming{\textsc{lemming}}

\renewcommand{\vec}[1]{\mathbf{#1}}

\aclfinalcopy 
\def\aclpaperid{154} 

\title{A Structured Variational Autoencoder\\for Contextual Morphological Inflection}

\author{
  Lawrence Wolf-Sonkin$^*$\; Jason Naradowsky$^*$ \; Sabrina J. Mielke$^*$ \; Ryan Cotterell\thanks{\; All authors contributed equally.}\\
  Department of Computer Science, Johns Hopkins University\\
  {\tt \{lawrencews,narad,sjmielke,ryan.cotterell\}@jhu.edu}  \\}

\date{}

\begin{document}

\thispagestyle{plain}
\pagestyle{plain}

\maketitle

\begin{abstract}
  Statistical morphological inflectors are typically trained on fully
  supervised, type-level data. One remaining open research question is
  the following: How can we effectively exploit raw, token-level data
  to improve their performance? To this end, we introduce a novel
  generative latent-variable  model for the semi-supervised
  learning of inflection generation. To enable posterior inference
  over the latent variables, we derive an efficient variational
  inference procedure based on the wake-sleep algorithm. We experiment
  on 23 languages, using the Universal Dependencies corpora in a simulated
  low-resource setting, and find improvements of over 10\% absolute accuracy in some cases.
\end{abstract}

\section{Introduction}
The majority of the world's languages overtly encodes syntactic
information on the word form itself, a phenomenon termed inflectional
morphology \cite{dryer2005world}. In English, for example, the verbal
lexeme with lemma \word{talk} has the four forms: \word{talk}, \word{talks},
\word{talked} and \word{talking}.  Other languages, such as Archi
\cite{kibrik1998archi}, distinguish more than a thousand verbal
forms. Despite the cornucopia of unique variants a single lexeme may
mutate into, native speakers can flawlessly predict the correct variant
that the lexeme's syntactic context dictates. Thus,
in computational linguistics, a natural question is the following: Can
we estimate a probability model that can do the same?

The topic of inflection generation has been the focus of a flurry of
individual attention of late
and, moreover, has been the subject of two shared tasks
\cite{cotterell-EtAl:2016:SIGMORPHON,cotterell-conll-sigmorphon2017}. Most
work, however, has focused on the fully supervised case---a source
lemma and the morpho-syntactic properties are fed into a model, which
is asked to produce the desired inflection. In contrast, our work
focuses on the semi-supervised case, where we wish to
make use of unannotated raw text, i.e., a sequence of inflected tokens.

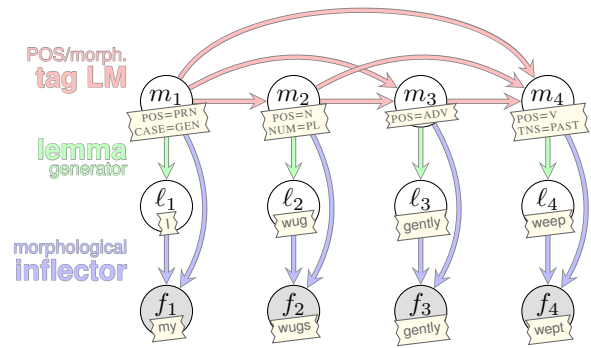
\begin{figure}
  \begin{adjustbox}{width=1.\columnwidth}
    \tikzset{
      double arrow/.style args={#1 colored by #2 and #3}{-stealth,line width=#1,#2, 
        postaction={draw,-stealth,#3,line width=2*(#1)/3,shorten <=(#1)/15,shorten >=(#1)/3}, 
      }
    }%
    \tikzstyle{arrow} = [->, >={triangle 45}]%
    \tikzstyle{red_arrow} = [double arrow=2pt colored by gray and cbcol1]%
    \tikzstyle{green_arrow} = [double arrow=2pt colored by gray and cbcol2]%
    \tikzstyle{blue_arrow} = [double arrow=2pt colored by gray and cbcol3]%
    \tikzstyle{exampleword} = [draw=black!60, fill=yellow!10!white, scale=0.6, rotate=7]%
    \makeatletter%
    \pgfdeclareshape{papersnip}{%
    \inheritsavedanchors[from=rectangle]%
    \inheritanchor[from=rectangle]{center}%
    \inheritanchor[from=rectangle]{mid}%
    \inheritanchor[from=rectangle]{base}%
    \inheritanchor[from=rectangle]{north}%
    \inheritanchor[from=rectangle]{south}%
    \backgroundpath{%
    \tikz@options%
    \southwest\pgf@xa=\pgf@x \pgf@ya=\pgf@y%
    \northeast\pgf@xb=\pgf@x \pgf@yb=\pgf@y%
    \pgfpointdiff{\southwest}{\northeast}\pgf@xc=\pgf@x \pgf@yc=\pgf@y%
    \pgfpathmoveto{\pgfpoint{\pgf@xa - 1}{\pgf@ya}}%
    \pgfpathlineto{\pgfpoint{\pgf@xa + 1}{0.75\pgf@ya + 0.25\pgf@yb}}%
    \pgfpathlineto{\pgfpoint{\pgf@xa - 1}{0.50\pgf@ya + 0.50\pgf@yb}}%
    \pgfpathlineto{\pgfpoint{\pgf@xa + 1}{0.25\pgf@ya + 0.75\pgf@yb}}%
    \pgfpathlineto{\pgfpoint{\pgf@xa - 1}{\pgf@yb}}%
    \pgfpathlineto{\pgfpoint{\pgf@xb + 1}{\pgf@yb}}%
    \pgfpathlineto{\pgfpoint{\pgf@xb - 1}{0.75\pgf@yb + 0.25\pgf@ya}}%
    \pgfpathlineto{\pgfpoint{\pgf@xb + 1}{0.50\pgf@yb + 0.50\pgf@ya}}%
    \pgfpathlineto{\pgfpoint{\pgf@xb - 1}{0.25\pgf@yb + 0.75\pgf@ya}}%
    \pgfpathlineto{\pgfpoint{\pgf@xb + 1}{\pgf@ya}}%
    \pgfpathclose%
    }%
    }%
    \makeatother%
    \hspace*{-1.5em}
    \begin{tikzpicture}
    \tikzstyle{connect}=[-latex, thick]
    
      \node[latent] (m1) {\smash{\raisebox{0em}{$m_1$}}};
      \node[latent] (m2) [right=of m1] {\smash{\raisebox{0em}{$m_2$}}};
      \node[latent] (m3) [right=of m2] {\smash{\raisebox{0em}{$m_3$}}};
      \node[latent] (m4) [right=of m3] {\smash{\raisebox{0em}{$m_4$}}};
    
      \node[latent] (l1) [below=of m1, yshift=.75em] {\smash{\raisebox{0em}{$\ell_1$}}};
      \node[latent] (l2) [below=of m2, yshift=.75em] {\smash{\raisebox{0em}{$\ell_2$}}};
      \node[latent] (l3) [below=of m3, yshift=.75em] {\smash{\raisebox{0em}{$\ell_3$}}};
      \node[latent] (l4) [below=of m4, yshift=.75em] {\smash{\raisebox{0em}{$\ell_4$}}};
    
      \node[obs] (f1) [below=of l1, yshift=.75em] {\smash{\raisebox{0em}{$f_1$}}};
      \node[obs] (f2) [below=of l2, yshift=.75em] {\smash{\raisebox{0em}{$f_2$}}};
      \node[obs] (f3) [below=of l3, yshift=.75em] {\smash{\raisebox{0em}{$f_3$}}};
      \node[obs] (f4) [below=of l4, yshift=.75em] {\smash{\raisebox{0em}{$f_4$}}};
    
      \draw [style=green_arrow] (m1) to (l1);
      \draw [style=blue_arrow] (l1) to (f1);
      \draw [style=blue_arrow, bend left, looseness=1.00] (m1) to (f1);
    
      \draw [style=green_arrow] (m2) to (l2);
      \draw [style=blue_arrow] (l2) to (f2);
      \draw [style=blue_arrow, bend left, looseness=1.00] (m2) to (f2);
    
      \draw [style=green_arrow] (m3) to (l3);
      \draw [style=blue_arrow] (l3) to (f3);
      \draw [style=blue_arrow, bend left, looseness=1.00] (m3) to (f3);
    
      \draw [style=green_arrow] (m4) to (l4);
      \draw [style=blue_arrow] (l4) to (f4);
      \draw [style=blue_arrow, bend left, looseness=1.00] (m4) to (f4);
    
      \draw [style=red_arrow] (m1) to (m2);

      \draw [style=red_arrow] (m2) to (m3);
      \draw [style=red_arrow, bend left, looseness=1.00] (m1) to (m3);
    
      \draw [style=red_arrow] (m3) to (m4);
      \draw [style=red_arrow, bend left, looseness=1.00] (m2) to (m4);
      \draw [style=red_arrow, bend left=60, looseness=0.75] (m1) to (m4);
    
      \contourlength{.02em}
      \node[left=of m1, yshift=1.6em, xshift=2.50em]{\color{cbcol1}\contour{gray!90}{\textsf{\scriptsize POS/morph.}}};
      \node[left=of m1, yshift=.75em, xshift=2.50em]{\color{cbcol1}\contour{gray!90}{\textsf{\bfseries tag LM}}};
    
      \node[left=of m1, yshift=-1.7em, xshift=2.50em]{\color{cbcol2}\contour{gray!90}{\textsf{\bfseries lemma}}};
      \node[left=of m1, yshift=-2.45em, xshift=2.50em]{\color{cbcol2}\contour{gray!90}{\textsf{\scriptsize generator}}};
    
      \node[left=of l1, yshift=-1.5em, xshift=2.50em]{\color{cbcol3}\contour{gray!90}{\textsf{\scriptsize morphological}}};
      \node[left=of l1, yshift=-2.25em, xshift=2.50em]{\color{cbcol3}\contour{gray!90}{\textsf{\bfseries inflector}}};
    
      \node[papersnip, overlay, below=of m1, yshift=3.3em, style=exampleword, align=center, inner sep=0em] {
        \color{black!60}\footnotesize
        \hspace*{-.7em}
        \begin{tabular}{rcl}
          \slot{pos} & \hspace*{-1.3em}\textsc{=}\hspace*{-1.3em} & \pos{prn}\\[-.2em]
          \slot{case} & \hspace*{-1.3em}\textsc{=}\hspace*{-1.3em} & \slot{gen}\\
        \end{tabular}
        \hspace*{-.7em}
      };
      \node[papersnip, overlay, below=of m2, yshift=3.3em, style=exampleword, align=center, inner sep=0em] {
        \color{black!60}\footnotesize
        \hspace*{-.7em}
        \begin{tabular}{rcl}
          \slot{pos} & \hspace*{-1.3em}\textsc{=}\hspace*{-1.3em} & \pos{n}\\[-.2em]
          \slot{num} & \hspace*{-1.3em}\textsc{=}\hspace*{-1.3em} & \slot{pl}\\
        \end{tabular}
        \hspace*{-.7em}
      };
      \node[papersnip, overlay, below=of m3, yshift=3.3em, style=exampleword, align=center, inner sep=0em] {
        \color{black!60}\footnotesize
        \hspace*{-.7em}
        \begin{tabular}{rcl}
          \slot{pos} & \hspace*{-1.3em}\textsc{=}\hspace*{-1.3em} & \pos{adv}\\
        \end{tabular}
        \hspace*{-.7em}
      };
      \node[papersnip, overlay, below=of m4, yshift=3.3em, style=exampleword, align=center, inner sep=0em] {
        \color{black!60}\footnotesize
        \hspace*{-.7em}
        \begin{tabular}{rcl}
          \slot{pos} & \hspace*{-1.3em}\textsc{=}\hspace*{-1.3em} & \pos{v}\\[-.2em]
          \slot{tns} & \hspace*{-1.3em}\textsc{=}\hspace*{-1.3em} & \slot{past}\\
        \end{tabular}
        \hspace*{-.7em}
      };
    
      \node[papersnip, overlay, below=of l1, yshift=3.3em, style=exampleword] {\color{black!60} \word{I}};
      \node[papersnip, overlay, below=of l2, yshift=3.3em, style=exampleword] {\color{black!60} \word{wug}};
      \node[papersnip, overlay, below=of l3, yshift=3.3em, style=exampleword] {\color{black!60} \word{gently}};
      \node[papersnip, overlay, below=of l4, yshift=3.3em, style=exampleword] {\color{black!60} \word{weep}};
    
      \node[papersnip, overlay, below=of f1, yshift=3.3em, style=exampleword] {\color{black!60} \word{my}};
      \node[papersnip, overlay, below=of f2, yshift=3.3em, style=exampleword] {\color{black!60} \word{wugs}};
      \node[papersnip, overlay, below=of f3, yshift=3.3em, style=exampleword] {\color{black!60} \word{gently}};
      \node[papersnip, overlay, below=of f4, yshift=3.3em, style=exampleword] {\color{black!60} \word{wept}};
    
    \end{tikzpicture}
    
    \end{adjustbox}
  \caption{A length-$4$ example of our generative model factorized as in \cref{eq:joint} and overlayed with example values of the random variables in the sequence. We highlight that all the conditionals
    in the Bayesian network are recurrent neural networks, e.g., we note that $m_i$ depends on
  $\vm_{< i}$ because we employ a recurrent neural network to model the morphological tag sequence. }
  \label{fig:model}
\end{figure}

Concretely, we develop a generative directed graphical model of
inflected forms \emph{in context}.  A
contextual inflection model works as follows: Rather than just
generating the proper inflection for a single given word form \emph{out of context} (for example
\textit{walking} as the gerund of \textit{walk}), our generative model
is actually a fully-fledged language model. In other words, it generates
\emph{sequences} of inflected words. The graphical model is displayed in \cref{fig:model}
and examples of words it may generate are pasted on top of the graphical model notation.
That our model is a language model enables it to exploit both inflected lexicons and unlabeled raw
text in a principled semi-supervised way. In order to train using raw-text corpora
(which is useful when we have less annotated data), we marginalize out
the unobserved lemmata and morpho-syntactic annotation from unlabeled
data. In terms of \cref{fig:model}, this refers to marginalizing out $m_1, \ldots, m_4$
and $\ell_1, \ldots, \ell_4$. 
As this marginalization is intractable, we derive a variational
inference procedure that allows for efficient approximate
inference. Specifically, we modify the wake-sleep procedure of
\newcite{hinton1995wake}. It is the inclusion of raw
text in this fashion that makes our model \emph{token level}, a novelty in the camp of
inflection generation, as much recent work in inflection
generation \cite{dreyer-smith-eisner:2008:EMNLP,durrett2013supervised,nicolai2015inflection,ahlbergforsberg2015,faruqui2015morphological},
trains a model on \emph{type-level} lexicons.

We offer empirical validation of our
model's utility with experiments on 23 languages from the
Universal Dependencies corpus in a simulated low-resource setting.\footnote{We make our code and data available at: \url{https://github.com/lwolfsonkin/morph-svae}.}
Our semi-supervised scheme improves inflection generation by
over 10\% absolute accuracy in some cases.

\begin{table}
\centering
\begin{tabular}{lllll}
  \toprule
  & {\sc sg} & {\sc pl} & {\sc sg} & {\sc pl}  \\ \midrule
      {\sc nom} & \word{Wort} & \word{W{\"o}rter} & \word{Herr} & \word{Herren} \\
    {\sc gen} & \word{Wortes} & \word{W{\"o}rter} & \word{Herrn} & \word{Herren} \\
{\sc acc} &  \word{Wort} & \word{W{\"o}rter}  & \word{Herrn} & \word{Herren} \\
{\sc dat}  & \word{Worte} & \word{W{\"o}rtern} & \word{Herrn} & \word{Herren} \\
\bottomrule
\end{tabular}
\caption{As an exhibit of morphological inflection, full paradigms (two numbers and four cases, 8 slots total) for the German nouns \scalebox{0.85}{\word{Wort}} (``word'') and \scalebox{0.85}{\word{Herr}} (``gentleman''), with abbreviated and tabularized UniMorph annotation. 
}
\label{tab-paradigm}
\end{table}

\section{Background: Morphological Inflection}\label{sec:inflectional-morphology}

\subsection{Inflectional Morphology}

To properly discuss models of inflectional morphology, we require a formalization.
We adopt the framework of word-based morphology
\cite{aronoff1976word,spencer1991morphological}. Note
in the present paper, we omit derivational morphology.

We define an \defn{inflected lexicon} as a set of
4-tuples consisting of a part-of-speech tag, a lexeme, an
inflectional slot, and a surface form.
A \defn{lexeme} is a discrete object
that indexes the word's
core meaning and part of speech.
In place of such an abstract lexeme, lexicographers will often use a \defn{lemma}, denoted by $\ell$, which is a designated\footnote{A specific slot of the paradigm is chosen, depending on the part-of-speech tag -- all these terms are defined next.} surface form
of the lexeme (such as the infinitive). For the remainder
of this paper, we will use the lemma as a proxy for the lexeme, wherever convenient,
although we note that lemmata may be ambiguous:
\word{bank} is the lemma for at least two distinct nouns
and two distinct verbs. For inflection, this ambiguity will
rarely\footnote{One example of a paradigm where the lexeme, rather
  than the lemma, may influence inflection is
  \scalebox{0.9}{\word{hang}.} If one chooses the lexeme that
  licenses animate objects, the proper past tense is \scalebox{0.9}{\word{hanged}},
  whereas it is \scalebox{0.9}{\word{hung}} for the lexeme that licenses inanimate
  objects.}
play a role---for instance, all senses of
\word{bank} inflect in the same fashion.

A \defn{part-of-speech (POS) tag},
denoted $t$, is a coarse syntactic category such as \pos{Verb}.
Each POS tag allows some set of lexemes, and also allows some set of
inflectional \defn{slots}, denoted as $\sigma$, such as $\brackets{\attval{tns}{past},\attval{person}{3}}$.
Each allowed $\angles{\text{tag, lexeme, slot}}$
triple is realized---in only one way---as an inflected \defn{surface
form}, a string over a fixed phonological or orthographic alphabet
$\Sigma$.  (In this work, we take $\Sigma$ to be an orthographic
alphabet.) Additionally, we will define the term \defn{morphological
tag}, denoted by $m$, which we take to be the POS-slot
pair $m = \langle t, \sigma \rangle$. We will
further define ${\cal T}$ as the set of all
POS tags and ${\cal M}$ as the set of all morphological tags.

A \defn{paradigm} $\pi(t, \ell)$ is the mapping from tag $t$'s slots
to the surface forms that ``fill'' those slots for lexeme/lemma $\ell$.  For
example, in the English paradigm $\pi(\pos{Verb}, \word{talk})$, the
past-tense slot is said to be filled by \word{talked}, meaning that
the lexicon contains the tuple $\angles{\pos{Verb}, \word{talk},
  \slot{past}, \word{talked}}$.

A cheat sheet for the notation is provided in \cref{tab:notation}.

We will specifically work with the UniMorph annotation scheme
\cite{sylak2016composition}. Here, each slot specifies a
morpho-syntactic bundle of inflectional features such as tense, mood, person, number,
and gender.  For example, the German surface form \word{W{\"o}rtern}
is listed in the lexicon with tag \pos{Noun}, lemma \word{Wort}, and a slot specifying the
feature bundle $\brackets{\attval{num}{pl}, \attval{case}{dat}}$.
The full paradigms $\pi(\pos{Noun},\word{Wort})$ and $\pi(\pos{Noun},\word{Herr})$ are found in \cref{tab-paradigm}.

\begin{table}
  \centering
  \begin{adjustbox}{width=1.\columnwidth}
  \begin{tabular}{lll} \toprule
    \textbf{object} & \textbf{symbol} & \textbf{example} \\ \midrule 
    form & $f$ & \word{talking} \\
    lemma & $\ell$ & \word{talk} \\
    POS & $t$ & \pos{Verb} \\
    slot & $\sigma$ & $\brackets{\attval{tns}{gerund}}$ \\
    morph. tag & $m$ & $\brackets{\attval{pos}{v},\attval{tns}{gerund}}$ \\ \bottomrule
  \end{tabular}
  \end{adjustbox}
  \caption{Notational cheat sheet for the paper.}
  \label{tab:notation}
\end{table}

\subsection{Morphological Inflection}\label{sec:inflection-notation}
Now, we formulate the task of context-free \textbf{morphological inflection} using the
notation developed in \cref{sec:inflectional-morphology}. Given a set
of $N$ form-tag-lemma triples $\{\langle f_i, m_i, \ell_i \rangle \}_{i=1}^N$,
the goal of morphological inflection is to map
the pair $\langle m_i, \ell_i \rangle$ to the form $f_i$. As the
definition above indicates, the task is traditionally performed at the
\emph{type level}. In this work, however, we focus on a generalization of the
task to the \emph{token level}---we seek to map a bisequence of lemma-tag pairs to the sequence of inflected
forms in context. Formally, we will denote the lemma-morphological tag bisequence as $\langle \vell, \vm \rangle$
and the form sequence as $\vf$. Foreshadowing, the primary motivation for this generalization is to enable the use
of raw-text in a semi-supervised setting. 

\section{Generating Sequences of Inflections}\label{sec:generative}
The primary contribution
of this paper is a novel generative
model over sequences of inflected words in their sentential context.
Following the notation
laid out in \cref{sec:inflection-notation}, we seek
to jointly learn a distribution over sequences of forms $\vf$, lemmata $\vell$, and morphological tags $\vm$.
The generative procedure is as follows:
First, we sample a sequence of tags $\vm$,
each morphological tag coming from a
language model over morphological tags:  $m_i \sim \ptheta(\cdot \mid \vm_{< i})$.
Next, we sample the sequence of lemmata $\vl$
given the previously sampled sequence of tags $\vm$---
these are sampled conditioned only on the corresponding morphological tag: $\ell_i \sim \ptheta(\cdot \mid m_i)$. 
Finally, we sample the sequence of inflected words $\vf$,
where, again, each word is chosen conditionally
independent of other elements of the sequence: $f_i \sim \ptheta(\cdot \mid \ell_i, m_i)$.%
\footnote{Note that we denote all three distributions as $\ptheta$ to simplify notation and emphasize the joint modeling aspect; context will always resolve the ambiguity in this paper. We will discuss their parameterization in \cref{sec:parameterization}.}
This yields the factorized joint distribution:
\begin{align}\label{eq:joint}
  \ptheta&(\vf, \vl, \vm) = \\ 
  & \Bigg( \prod_{i=1}^{|\vf|} \underbrace{\ptheta(f_i \mid \ell_i, m_i)}_{\substack{\textit{morphological inflector} \\ \numberingBlueB{3} }} \!\!\cdot \underbrace{\ptheta(\ell_i \mid m_i)}_{\substack{\textit{lemma generator} \\  \numberingGreenB{2}}} \Bigg) \!\cdot \underbrace{\ptheta(\vm)}_{\substack{\textit{m-tag LM} \\  \numberingRedB{1}}}  \nonumber 
\end{align}
We depict the corresponding directed graphical model in \cref{fig:model}. 

\paragraph{Relation to Other Models in NLP.}
As the graphical model drawn in \cref{fig:model} shows, our model is
quite similar to a Hidden Markov Model (HMM) \cite{rabiner1989tutorial}. There are two primary
differences. First, we remark that an HMM directly emits a form $f_i$ conditioned on
the tag $m_i$. Our model, in contrast, emits a lemma
$\ell_i$ conditioned on the morphological tag $m_i$ and, then, conditioned on both the lemma
$\ell_i$ and the tag $m_i$, we emit the inflected form $f_i$. In this sense, our model
resembles the hierarchical HMM of \newcite{fine1998hierarchical} with the difference
that we do not have interdependence between the lemmata $\ell_i$. The second difference
is that our model is non-Markovian: we sample the $i^\text{th}$ morphological
tag $m_i$ from a distribution that depends on \emph{all} previous tags, using an LSTM language model (\cref{sec:lstm-lm}). This
yields richer interactions among the tags, which may be necessary for modeling
long-distance agreement phenomena.

\paragraph{Why a Generative Model?}
What is our interest in a generative model of inflected forms?
\cref{eq:joint} is a syntax-only language model in that
it only allows for interdependencies between the morpho-syntactic tags in
$\ptheta(\vm)$. However, given a tag sequence $\vm$, the individual lemmata
and forms are \emph{conditionally independent}. This prevents the
model from learning notions such as semantic frames and topicality.
So what is this model good for? Our chief interest is
the ability to \emph{train a morphological inflector on unlabeled data},
which is a boon in a low-resource setting.
As the model is generative, we may consider the
latent-variable model:
\begin{equation}
  \ptheta(\vf) = \sum_{\langle \vell, \vm \rangle} \ptheta(\vf, \vell, \vm), \label{eq:marginal}
\end{equation}
where we marginalize out the latent lemmata and morphological tags from
raw text. 
The sum in \cref{eq:marginal} is unabashedly intractable---given a
sequence $\vf$, it involves consideration
of an exponential (in $\size{\vf}$) number of tag sequences and an \emph{infinite} number
of lemmata sequences. Thus, we will fall back on an approximation scheme (see \cref{sec:inference}).

\section{Recurrent Neural Parameterization}\label{sec:parameterization}
The graphical model from \cref{sec:generative} specifies a
family of models that obey the conditional independence assumptions
dictated by the graph in \cref{fig:model}. In this section we define a
specific parameterization using long short-term memory (LSTM) recurrent
neural network \cite{hochreiter1997long} language models \cite{sundermeyer2012lstm}.

\subsection{LSTM Language Models}\label{sec:lstm-lm}

Before proceeding, we review the modeling of sequences with LSTM language models. Given some
alphabet $\Delta$, the distribution over sequences $\vec{x} \in \Delta^*$
can be defined as follows:
\begin{equation}
  p(\vec{x}) = \prod_{j=1}^{|\vec{x}|} p(x_j \mid \vec{x}_{<j}),
\end{equation}
where $\vec{x}_{<j} = x_1, \ldots, x_{j-1}$.
The prediction at time step $j$ of a single element $x_j$ is then parametrized by a neural network:
\begin{equation}
  p(x_j \mid \vec{x}_{<j}) = \text{softmax}\left(\mathbf{W}\cdot\mathbf{h}_j+\mathbf{b} \right),
\end{equation}
where $\mathbf{W} \in \mathbb{R}^{\size{\Delta} \times d}$ and $\vec{b} \in \mathbb{R}^{\size{\Delta}}$ are learned parameters (for some number of hidden units $d$) and the hidden state $\vec{h}_j \in \mathbb{R}^d$ is defined through the recurrence given by \citet{hochreiter1997long} from the previous hidden state and an embedding of the previous character (assuming some learned embedding function $\mathbf{e}\colon \Delta \to \mathbb{R}^c$ for some number of dimensions $c$):
\begin{equation}
  \vec{h}_j = \mathrm{LSTM}\big( \, \vec{h}_{j-1},\, \mathbf{e}(x_{j-1})  \, \big)
\end{equation}

\subsection{Our Conditional Distributions}
We discuss each of the
factors in \cref{eq:joint} in turn.
\paragraph{\numberingRed{1} Morphological Tag Language Model: $\ptheta(\vm)$.}
We define $\ptheta(\vm)$ as an LSTM language model, as
defined in \cref{sec:lstm-lm}, where we take $\Delta = {\cal M}$, i.e., the elements of the sequence that are to be predicted are tags like $\brackets{\attval{pos}{v},\attval{tns}{gerund}}$. Note that the embedding function $\mathbf{e}$ does not treat them as atomic units, but breaks them up into individual attribute-value pairs that are embedded individually and then summed to yield the final vector representation. To be precise, each tag is first encoded by a multi-hot vector, where
each component corresponds to a attribute-value pair in the slot, and then this multi-hot vector is multiplied with an embedding matrix.

\paragraph{\numberingGreen{2} Lemma Generator: $\ptheta(\ell_i \mid m_i)$.}
The next distribution in our model is a lemma generator which we define to be a \textit{conditional} LSTM language model over characters (we take $\Delta = \Sigma$), i.e., each $x_i$ is a single (orthographic) character. The language model is conditioned on $t_i$ (the part-of-speech information contained in the morphological tag $m_i=\angles{t_i, \sigma_i}$), which we embed into a low-dimensional space and feed
to the LSTM by concatenating its embedding with that of the current character. Thusly, we obtain the new recurrence relation for the hidden state:
\begin{equation}
    \vec{h}_j = \mathrm{LSTM}\Big( \vec{h}_{j-1}, \Big[\,\mathbf{e}\big([\ell_i]_{j-1}\big) \,;\, \mathbf{e}'\big(t_i\big)\,\Big] \Big),
\end{equation}
where $[\ell_i]_j$ denotes the $j^\text{th}$ character of the generated lemma $\ell_i$ and
$\mathbf{e}': \mathcal{T} \to \mathbb{R}^{c'}$ for some $c'$ is a learned embedding function for POS tags. Note that we embed
only the POS tag, rather than the entire morphological tag, as we assume the lemma depends on the part of speech exclusively.

\paragraph{\numberingBlue{3} Morphological Inflector: $\ptheta(f_i \mid \ell_i, m_i)$.}
The final conditional in our model is a morphological inflector, which
we parameterize as a neural recurrent sequence-to-sequence model
\cite{DBLP:conf/nips/SutskeverVL14} with Luong dot-style attention
\cite{luong-pham-manning:2015:EMNLP}.
Our particular model uses a single encoder-decoder architecture \citep{kann-schutze:2016} for all tag pairs within a language and we refer to reader to that
paper for further details. 
Concretely, the encoder runs over a string consisting of the desired slot
and all characters of the lemma that is to be inflected
(e.g. \texttt{<w> V PST t a l k </w>}), one LSTM
running left-to-right, the other right-to-left.  Concatenating the
hidden states of both RNNs at each time step results in hidden states
$\vec{h}^{(\textit{enc})}_j$.  The decoder, again, takes the form of an LSTM
language model (we take $\Delta = \Sigma$), producing the inflected form character by character, but at each time step not only
the previous hidden state and the previously generated token are
considered, but attention (a convex combination)
over all encoder hidden states
$\vec{h}^{(\text{enc})}_j$, with the distribution given by another
neural network; see \newcite{luong-pham-manning:2015:EMNLP}.

\section{Semi-Supervised Wake-Sleep}\label{sec:inference}
We train the model with the wake-sleep
procedure, which requires us to perform posterior inference over
the latent variables. 
However, the exact computation in the model is
intractable---it involves a sum over all possible lemmatizations and
taggings of the sentence, as shown in \cref{eq:marginal}. Thus, we fall back
on a variational approximation
\cite{Jordan:1999:IVM:339248.339252}. We train an \textbf{inference
  network} $\qphi(\vell, \vm \mid \vf)$ that approximates the
true posterior over the latent variables $\ptheta(\vell, \vm \mid
\vf)$.\footnote{Inference networks are also known as
  \textbf{stochastic inverses} \cite{stuhlmuller2013learning} or \textbf{recognition models} \cite{dayan1995helmholtz}.}
The variational family we choose in this work will be detailed in \cref{sub:our_variational_family}.
We fit the distribution $\qphi$ using a semi-supervised
extension of the wake-sleep algorithm
\cite{hinton1995wake,dayan1995helmholtz,DBLP:journals/corr/BornscheinB14}. We
derive the algorithm in the following subsections and provide pseudo-code in
\cref{alg:wake-sleep}.

Note that the wake-sleep algorithm shows
structural similarities to the expectation-maximization (EM) algorithm \citep{DemLaiRub77Maximum},
and, presaging the exposition, we note that the wake-sleep procedure
is a type of variational EM \cite{beal2003variational}. The key
difference is that the E-step minimizes an inclusive KL divergence,
rather than the exclusive one typically found in variational EM.

\subsection{Data Requirements of Wake-Sleep}

We emphasize again that we will train
our model in a semi-supervised fashion. Thus,
we will assume a set of labeled sentences,
$\Dlabeled$, represented as a set of triples $\angles{\vf, \vell, \vm}$, and a
set of unlabeled sentences, $\Dunlabeled$, represented as a set of surface form sequences $\vf$.

\subsection{The Sleep Phase}
Wake-sleep first
dictates that we find an approximate posterior distribution $\qphi$
that minimizes the KL divergences for all form sequences:
\begin{equation}
 D_\textit{KL}\Big( \underbrace{\ptheta(\cdot, \cdot, \cdot)}_{\textit{full joint: $\cref{eq:joint}$}} \mid\mid \underbrace{\qphi(\cdot, \cdot \mid \cdot)}_{\textit{variational approximation}} \Big) \label{eq:kl}
\end{equation}
with respect to the parameters $\vphi$, which control the variational
approximation $\qphi$. Because $\qphi$ is trained to be a variational
approximation for \emph{any} input $\vf$, it is called an inference network.
In other words, it
will return an approximate posterior over the latent variables for any observed
sequence.
Importantly, note that computation of
\cref{eq:kl} is still hard---it requires us to normalize the
distribution $\ptheta$, which, in turn, involves a sum
over all lemmatizations and taggings. However, it does lend itself to
an efficient Monte Carlo approximation. As our model is fully
generative and directed, we may easily take samples from the complete
joint. Specifically, we will take $K$ samples $\angles{\tilvf, \tilvell, \tilvm}
\sim \ptheta(\cdot, \cdot, \cdot)$ by forward sampling and define them as $\Ddreamt$. We remark that we
use a tilde to indicate that a form, lemmata or tag is
sampled, rather than human annotated. Using $K$ 
samples, we obtain the objective
\begin{equation}\label{eq:unsupervised-sleep}
{\cal S}_\textit{unsup} = \sfrac{1}{K} \cdot \mkern-35mu \sum_{\angles{\tilvf, \tilvell, \tilvm} \in \Ddreamt } \mkern-25mu \log
\qphi(\tilvell, \tilvm \mid
\tilvf),
\end{equation}
which we could maximize by fitting the
model $\qphi$ through backpropagation \cite{rumelhart1985learning},
as one would during maximum likelihood estimation.

\subsection{The Wake Phase}
Now, given our approximate posterior $\qphi(\vell, \vm \mid \vf)$, we are in a position to re-estimate
the parameters of the generative model $\ptheta(\vf, \vell, \vm)$. Given
a set of unannotated sentences $\Dunlabeled$, we again first consider the objective
\begin{equation}\label{eq:unsupervised-wake}
{\cal W}_\textit{unsup} = \sfrac{1}{M} \cdot \mkern-35mu \sum_{\angles{\vf, \tilvell, \tilvm} \in \Dwake} \mkern-25mu \log \ptheta(\vf, \tilvell, \tilvm)
\end{equation}
where $\Dwake$ is a set of triples $\angles{\vf, \tilvell, \tilvm}$ with $\vf \in \Dunlabeled$ and $\langle \tilvell, \tilvm \rangle \sim \qphi(\cdot, \cdot \mid \vf)$, maximizing with respect to the parameters $\vtheta$
(we may stochastically backprop through the expectation simply by backpropagating through this sum).
Note that \cref{eq:unsupervised-wake} is a Monte Carlo approximation of the inclusive divergence of the data distribution of $\Dunlabeled$ times $\qphi$ with $\ptheta$.

\subsection{Adding Supervision to Wake-Sleep}

So far we presented a purely unsupervised training method that makes no assumptions about the latent lemmata and morphological tags.
In our case, however, we have a very clear idea what the latent variables
should look like. For instance, we are quite certain that the lemma of
\word{talking} is \word{talk} and that it is in fact a {\sc gerund}.  And, indeed, we have access to
annotated examples $\Dlabeled$ in the form of an annotated
corpus. In the presence of these data, we optimize the supervised sleep phase objective,
\begin{equation}\label{eq:supervised-sleep}
{\cal S}_\textit{sup} = \sfrac{1}{N} \cdot \mkern-38mu \sum_{\angles{\vf, \vell, \vm} \in \Dlabeled} \mkern-25mu \log
\qphi(\vell, \vm \mid \vf).
\end{equation}
which is a Monte Carlo approximation of $D_\textit{KL}( \Dlabeled \mid\mid \qphi)$.
Thus, when fitting our variational approximation $\qphi$, we will
optimize a joint objective ${\cal S} = {\cal S}_\textit{sup} +
\gamma_\textit{sleep} \cdot {\cal S}_\textit{unsup}$, where ${\cal
  S}_\textit{sup}$, to repeat, uses actual annotated lemmata and morphological
tags; we balance the two parts of the objective with a scaling
parameter $\gamma_{\textit{sleep}}$.
Note that on the first sleep phase iteration,
we set $\gamma_\textit{sleep} = 0$ since taking samples
from an untrained $\ptheta(\cdot, \cdot, \cdot)$ when
we have available labeled data is of little utility.
We will discuss the provenance
of our data in \cref{sec:data}.

Likewise, in the wake phase we can neglect the approximation
$\qphi$ in favor of the annotated latent variables
found in $\Dlabeled$;
this leads
to the following supervised objective
\begin{equation}\label{eq:supervised-wake}
{\cal W}_\textit{sup} = \sfrac{1}{N} \cdot \mkern-38mu \sum_{\angles{\vf, \vell, \vm} \in \Dlabeled} \mkern-26mu \log \ptheta(\vf, \vell, \vm), 
\end{equation}
which is a Monte Carlo approximation of $D_\textit{KL}( \Dlabeled \mid\mid \ptheta)$.
As in the sleep phase, we will maximize ${\cal W} = {\cal W}_\textit{sup} + \gamma_\textit{wake} \cdot {\cal W}_\textit{unsup}$, where 
$\gamma_{\textit{wake}}$ is, again, a scaling parameter.

\begin{algorithm}[t]
  \begin{algorithmic}[1]
    \State \textbf{input} $\Dlabeled$ \Comment{\textit{labeled training data}}
    \State \textbf{input} $\Dunlabeled$ \Comment{\textit{unlabeled training data}}
    \For{$i=1$ \textbf{to} $I$}
      \State $\Ddreamt \gets \emptyset$
      \If{$i > 1$}
        \For{$k=1$ \textbf{to} $K$}
          \State $\langle\tilvf, \tilvell, \tilvm \rangle \sim \ptheta(\cdot, \cdot, \cdot)$
          \State $\Ddreamt \gets \Ddreamt \cup \{\langle \tilvf, \tilvell, \tilvm\rangle\}$
        \EndFor
      \EndIf
      \State maximize $\log \qphi$ on $\Dlabeled\,\cup\,\Ddreamt\qquad$\linebreak\hspace*{0em} \Comment{{\footnotesize \textit{this corresponds to \cref{eq:supervised-sleep} + \cref{eq:unsupervised-sleep}}}}
      \State $\Dwake \gets \emptyset$
      \For{$\vf \in \Dunlabeled$}
        \State $\langle \tilde{\vl}, \tilvm \rangle \sim \qphi\left(\cdot, \cdot \mid \vf \right)$
        \State $\Dwake \gets \Dwake \cup \{\langle \vf, \tilde{\vl}, \tilvm \rangle  \}$
      \EndFor
      \State maximize $\log \ptheta$ on $\Dlabeled\,\cup\,\Dwake\qquad$\linebreak\hspace*{0em} \Comment{{\footnotesize \textit{this corresponds to \cref{eq:supervised-wake} + \cref{eq:unsupervised-wake}}}}
    \EndFor
\end{algorithmic}
  \caption{semi-supervised wake-sleep}
  \label{alg:wake-sleep}
\end{algorithm}

\subsection{Our Variational Family}
\label{sub:our_variational_family}
How do we choose the variational family $\qphi$? In terms of NLP
nomenclature, $\qphi$ represents a joint morphological tagger and
lemmatizer.  The open-source tool \lemming{} \cite{muller-EtAl:2015:EMNLP} represents such an
object. \lemming{} is a higher-order
linear-chain conditional random field
\citep[CRF;][]{Lafferty:2001:CRF:645530.655813}, that is an extension of the
morphological tagger of
\newcite{mueller-schmid-schutze:2013:EMNLP}. Interestingly, \lemming{} is a linear model that makes use of simple character
$n$-gram feature templates. On both the tasks of morphological tagging
and lemmatization, neural models have supplanted linear models in
terms of performance in the high-resource case
\cite{heigold-neumann-vangenabith:2017:EACLlong}. However, we are
interested in producing an accurate approximation to the posterior in
the presence of minimal annotated examples and potentially noisy
samples produced during the sleep phase, where linear models still
outperform non-linear approaches
\cite{cotterell-heigold:2017:EMNLP2017}.  We note that
our variational approximation is compatible with any family.

\subsection{Interpretation as an Autoencoder}
We may also view our model as an autoencoder, following
\newcite{kingma2013auto}, who saw that a variational approximation to any
generative model naturally has this interpretation.
The crucial difference between \newcite{kingma2013auto} and this work is that our model is a \textbf{structured} variational autoencoder in the sense that the space of our latent code is structured: the inference network encodes a sentence into a pair of lemmata and
morphological tags $\angles{\vell, \vm}$. This bisequence is then decoded back into the
sequence of forms $\vf$ through a morphological inflector. The reason
the model is called an autoencoder is that we arrive
at an auto-encoding-like objective if we combine
the $\ptheta$ and $\qphi$ as so:
\begin{equation}\label{eq:auto-encoder}
  p(\vf \!\mid\!\hat{\vf})\!= \!\! \sum_{\langle \vl, \vm \rangle}
  \ptheta (\vf \!\mid\!  \vl, \vm) \cdot \qphi (\vl, \vm
  \!\mid\! \hat{\vf})
\end{equation}
where $\hat{\vf}$ is a copy of the original sentence $\vf$.

Note that this choice of latent space sadly precludes us from making
use of the \emph{reparametrization trick} that makes inference in VAEs
particularly efficient.
In fact, our whole inference procedure is quite different as we do not perform gradient descent on both $\qphi$ and $\ptheta$ jointly but alternatingly optimize both (using wake-sleep).
We nevertheless call our model a VAE to uphold the distinction between the VAE as a model (essentially a specific Helmholtz machine \cite{dayan1995helmholtz}, justified by variational inference) and the end-to-end inference procedure that is commonly used.

Another way of viewing this model is that it tries to force the words in the
corpus through a syntactic bottleneck. Spiritually, our work is
close to the conditional random field autoencoder of
\newcite{ammar2014conditional}.

We remark that many other structured NLP tasks can be ``autoencoded''
in this way and, thus, trained by a similar wake-sleep procedure. For
instance, any two tasks that effectively function as inverses, e.g.,
translation and backtranslation, or language generation and parsing,
can be treated with a similar variational autoencoder. While this work
only focuses on the creation of an improved morphological inflector
$\ptheta(\vf \!\mid \vell, \vm)$, one could imagine a situation where the encoder was also a
task of interest. That is, the goal would be to improve both the
decoder (the generation model) \emph{and} the encoder (the variational
approximation).

\section{Related Work}
Closest to our work is \newcite{zhou-neubig:2017:Long}, who describe
an unstructured variational autoencoder. However, the exact use case
of our respective models is distinct. Our method models the syntactic
dynamics with an LSTM language model over morphological tags. Thus, in
the semi-supervised setting, we require token-level
annotation. Additionally, our latent variables are interpretable as
they correspond to well-understood linguistic quantities. In contrast,
\newcite{zhou-neubig:2017:Long} infer latent lemmata as real vectors.
To the best of our knowledge, we are only the second attempt,
after \newcite{zhou-neubig:2017:Long}, to attempt to perform semi-supervised
learning for a neural inflection generator.  Other non-neural attempts
at semi-supervised learning of morphological inflectors include
\newcite{hulden-forsberg-ahlberg:2014:EACL}.  Models in this vein are
non-neural and often focus on exploiting corpus statistics, e.g.,
token frequency, rather than explicitly modeling the forms in
context. All of these approaches are designed to learn from a
type-level lexicon, rendering direct comparison difficult.

\begin{figure*}
    \begin{subfigure}[b]{0.64\columnwidth}
        \centering
        \includegraphics[width=1.15\columnwidth]{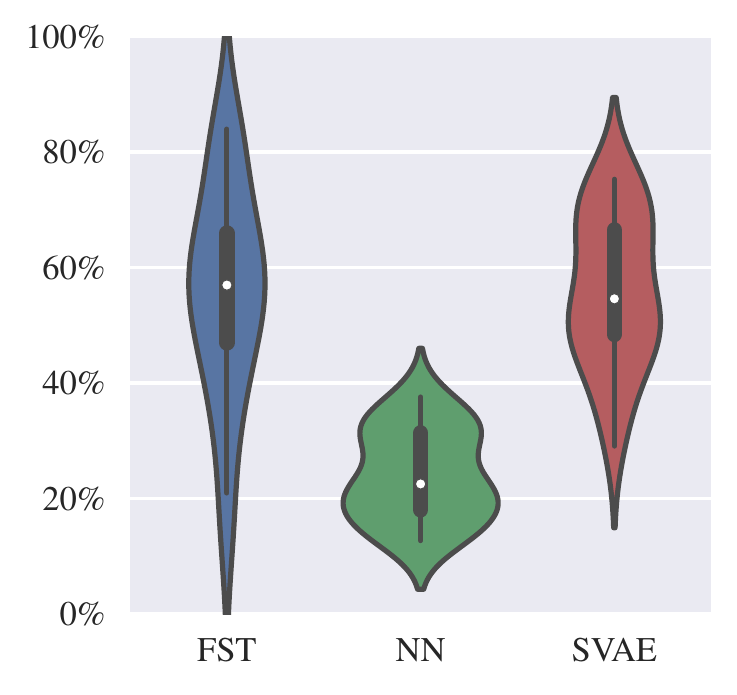}
        \caption{500 training tokens}
    \end{subfigure}
    ~ 
    \begin{subfigure}[b]{0.64\columnwidth}
        \centering
        \includegraphics[width=1.15\columnwidth]{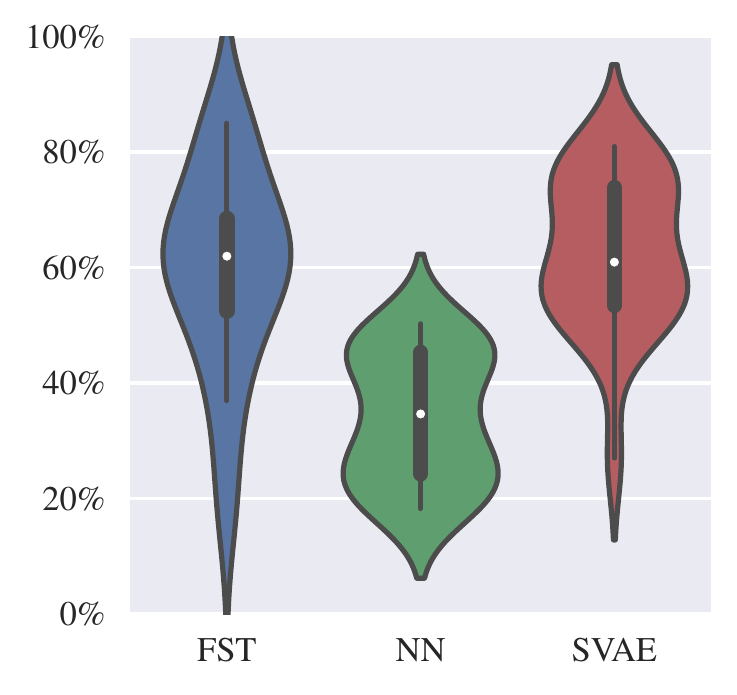}
        \caption{1000 training tokens}
    \end{subfigure}
    ~ 
    \begin{subfigure}[b]{0.64\columnwidth}
      \centering
      \includegraphics[width=1.15\columnwidth]{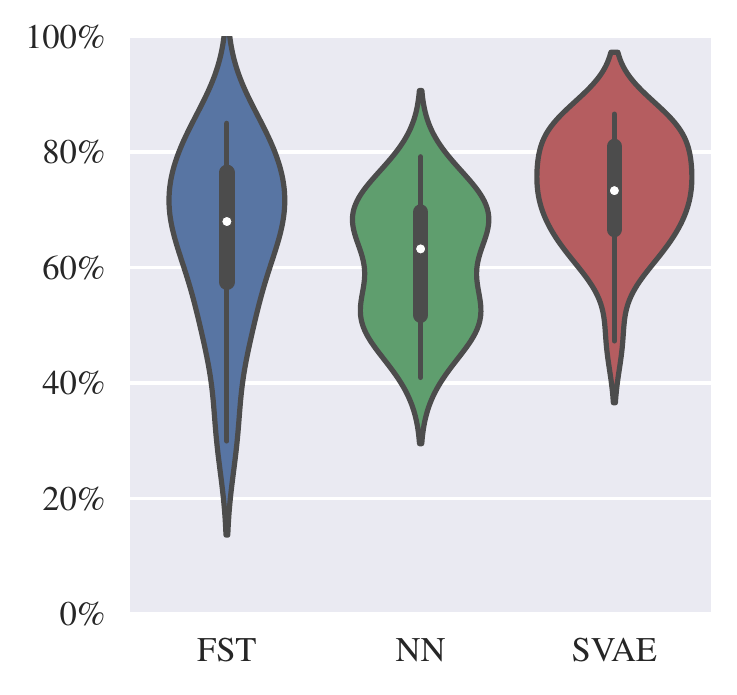}
      \caption{5000 training tokens}
    \end{subfigure}
    \vspace*{0.3cm}
    \caption{Violin plots showing the distribution over accuracies. The structured variational autoencoder (SVAE) always outperforms the neural network (NN), but only outperformed the FST-based approach when trained on 5000 annotated tokens. Thus, while semi-supervised training helps neural models reduce their sample complexity, roughly $5000$ annotated tokens are still required to boost their performance above more symbolic baselines. }
      \label{fig:violin}
\end{figure*}

\begin{table*}[t]
  \centering
\begin{adjustbox}{width=2.\columnwidth}
  \begin{tabular}{ccc | cccrr | cccrr | cccrr} \toprule
    &  & & \multicolumn{5}{c}{500 tokens} & \multicolumn{5}{c}{1000 tokens} & \multicolumn{5}{c}{5000 tokens} \\ \cmidrule(l){2-3} \cmidrule(l){4-8} \cmidrule(l){9-13} \cmidrule(l){14-18}
    & & \textbf{lang} & FST & NN & SVAE & $\vDelta_\textit{FST}$  & $\vDelta_\textit{NN}$ & FST & NN & SVAE & $\vDelta_\textit{FST}$  & $\vDelta_\textit{NN}$ & FST & NN & SVAE & $\vDelta_\textit{FST}$  & $\vDelta_\textit{NN}$\\ \midrule
    & \CA & ca & 81.0 & 28.11 & 71.76 & -9.24  & 43.65 & 85.0 & 42.58 & 78.46  & -6.54 & 35.88 & 84.0 & 74.22 & 85.77 & 1.77 & 11.55 \\
    & \FR & fr & 84.0 & 36.25 & 74.75 & -9.25  & 38.5 & 85.0 & 47.04 & 79.97  & -5.03 & 32.93 & 85.0 & 79.21 & 83.96 & -1.04 & 4.75 \\
    & \IT & it & 81.0 & 31.30 & 67.48 & -13.52  & 36.18 & 81.0 & 43.58 & 77.37  & -3.63 & 33.79 & 82.0 & 71.09 & 73.11 & -8.89 & 2.02 \\
    & \LA & la & 21.0 & 14.02 & 29.12 & 8.12  & 15.10 & 26.0 & 19.62 & 27.06  & 1.06 & 7.44 & 30.0 & 41.00 & 47.32 & 17.32 & 6.32 \\
   \rbox{Romance}  & \PT & pt & 81.0 & 31.58 & 72.54 & -8.46  & 40.96 & 83.0 & 47.27 & 73.24  & -9.76 & 25.97 & 82.0 & 74.17 & 86.13 & 4.13 & 11.96 \\
    & \RO & ro & 56.0 & 22.56 & 52.48 & -3.52  & 29.92 & 62.0 & 34.68 & 58.30  & -3.70 & 23.62 & 68.0 & 51.77 & 75.49 & 7.49 & 23.72 \\
    & \ES & es & 57.0 & 34.34 & 75.32 & 18.32  & 40.98 & 60.0 & 46.14 & 80.97  & 20.97 & 34.83 & 72.0 & 71.99 & 84.44 & 12.44 & 12.45 \\ \midrule
    & \NL & nl & 63.0 & 19.22 & 49.14 & -13.86  & 29.92 & 65.0 & 26.05 & 53.12  & -11.88 & 27.07 & 70.0 & 53.70 & 65.97 & -4.03 & 12.27 \\

    & \DA & da & 68.0 & 31.25 & 65.58 & -2.42  & 34.33 & 73.0 & 44.51 & 72.82  & -0.18 & 28.31 & 79.0 & 67.92 & 80.12 & 1.12 & 12.20 \\
    & \NO & no & 69.0 & 32.51 & 65.46 & -3.54  & 32.95 & 71.0 & 46.26 & 74.49  & 3.49 & 28.23 & 79.0 & 71.31 & 81.25 & 2.25 & 9.94 \\
    \rbox{Germanic}   & \NO & nn & 64.0 & 20.29 & 54.62 & -9.38  & 34.33 & 65.0 & 24.32 & 60.97  & -4.03 & 36.65 & 72.0 & 50.40 & 73.35 & 1.35 & 22.95 \\
    & \SV & sv & 63.0 & 19.02 & 58.15 & -4.85  & 39.13 & 66.0 & 36.35 & 67.18  & 1.18 & 30.83 & 74.0 & 59.82 & 78.23 & 4.23 & 18.41 \\ \midrule

    & \BG & bg & 44.0 & 15.51 & 47.22 & 3.22  & 31.71 & 51.0 & 21.00 & 57.18  & 6.18 & 36.18 & 59.0 & 49.06 & 71.15 & 12.15 & 22.09 \\
    & \PL & pl & 50.0 & 12.75 & 48.62 & -1.38  & 35.87 & 57.0 & 19.88 & 55.90  & -1.10 & 36.02 & 64.0 & 54.44 & 67.15 & 3.15 & 12.71 \\
    \rbox{\ \ \ Slavic}  & \SI & si & 52.0 & 15.60 & 55.69 & 3.69  & 40.09 & 61.0 & 26.39 & 61.22  & 0.22 & 34.83 & 68.0 & 66.65 & 75.40 & 7.40 & 8.75 \\
\midrule

    & \AR & ar & 14.0 & 31.47 & 63.53 & 49.53  & 32.06 & 17.0 & 48.53 & 71.52  & 54.52 & 22.99 & 34.0 & 68.16 & 80.72 & 46.72 & 12.56 \\
    \rbox{Semit.} & \HE & he & 60.0 & 37.61 & 71.11 & 11.11  & 33.50 & 66.0 & 50.28 & 76.32  & 10.32 & 26.04 & 72.0 & 64.37 & 86.60 & 14.6 & 22.23 \\ \midrule

    & \HU & hu & 53.0 & 22.56 & 48.64 & -4.36  & 26.08 & 56.0 & 28.62 & 60.74  & 4.74 & 32.12 & 61.0 & 66.45 & 72.84 & 11.84 & 6.39 \\
    & \ET & et & 39.0 & 21.81 & 42.16 & 3.16  & 20.35 & 45.0 & 29.66 & 51.75  & 6.75 & 22.09 & 49.0 & 46.82 & 58.91 & 9.91 & 12.09 \\
    \rbox{Finn.-Urg.} & \FI & fi & 37.0 & 12.97 & 35.78 & -1.22  & 22.81 & 42.0 & 19.03 & 47.65  & 5.65 & 28.62 & 49.0 & 46.75 & 62.76 & 13.76 & 16.01 \\ \midrule

    & \LV & lv & 57.0 & 17.16 & 48.29 & -8.71  & 31.13 & 63.0 & 18.30 & 53.58  & -9.42 & 35.28 & 66.0 & 51.84 & 66.12 & 0.12 & 14.28 \\
    & \EU & eu & 50.0 & 24.46 & 48.72 & -1.28  & 24.26 & 54.0 & 35.14 & 53.39  & -0.61 & 18.25 & 56.0 & 56.29 & 62.33 & 6.33 & 6.04 \\
    \rbox{\ \ \ \ other} & \TR & tr & 34.0 & 20.67 & 37.92 & 3.92  & 17.25 & 37.0 & 24.33 & 49.67  & 12.67 & 25.34 & 48.0 & 63.26 & 69.35 & 21.35 & 6.09 \\ \bottomrule
    & & avg & 55.57 & 24.04 & \textbf{55.83} & 0.26 & 31.79 & 59.61 & 33.89 & \textbf{62.73} & 3.12 & 6.90 & 65.35 & 60.90 & \textbf{73.41} & 8.06 & 12.51 \\
  \end{tabular}
\end{adjustbox}
  \caption{Type-level morphological inflection accuracy across different models, training scenarios, and languages}
  \label{tab:results}
\end{table*}

\section{Experiments}
While we estimate all the parameters in the generative model, the purpose of this work is to improve
the performance of morphological inflectors through semi-supervised learning with the incorporation of unlabeled data.

\subsection{Low-Resource Inflection Generation}
The development of our method was primarily aimed at the low-resource scenario, where we observe a limited
number of annotated data points. Why low-resource? When
we have access to a preponderance of data, morphological inflection is close to being a solved problem, as evinced
in SIGMORPHON's 2016 shared task. However, the CoNLL-SIGMORPHON 2017
shared task showed there is much progress to be made in the low-resource case. Semi-supervision is a clear avenue.

\subsection{Data}\label{sec:data}
As our model requires \emph{token-level} morphological annotation, we
perform our experiments on the Universal Dependencies (UD) dataset
\cite{11234/1-1983}.  As this stands in contrast to most
work on morphological inflection (which has used the UniMorph \cite{sylakglassman-EtAl:2015:ACL-IJCNLP}%
\footnote{The two annotation schemes are similar. For a discussion, we refer the reader to
  \url{http://universaldependencies.org/v2/features.html}; sadly there are differences that render all numbers
  reported in this work incomparable with previous work, see \cref{sec:baselines}.}
datasets), we use a converted version of UD data, in which the UD morphological tags have been deterministically converted into UniMorph tags.

For each of
the treebanks in the UD dataset, we divide the training portion into three
chunks consisting of the first 500, 1000 and 5000 tokens, respectively.
These \emph{labeled} chunks will constitute three unique sets $\Dlabeled$.  The remaining sentences in the
training portion will be used as \emph{unlabeled} data $\Dunlabeled$ for each
language, i.e., we will discard those labels. The development and test
portions will be left untouched.

\paragraph{Languages.}
We explore a typologically diverse set of languages of various stocks:
Indo-European, Afro-Asiatic, Turkic and Finno-Ugric, as well as the
language isolate Basque. We have organized our experimental languages
in \cref{tab:results} by genetic grouping, highlighting sub-families
where possible. The Indo-European languages mostly exhibit fusional
morphologies of varying degrees of complexity. The Basque, Turkic,
and Finno-Ugric languages are agglutinative. Both of the Afro-Asiatic
languages, Arabic and Hebrew, are Semitic and have templatic morphology
with fusional affixes.

\subsection{Evaluation}\label{sec:eval}
The end product of our procedure is a morphological inflector, whose performance
is to be improved through the incorporation of unlabeled data.  Thus, we evaluate
using the standard metric accuracy.
We will evaluate at the \emph{type level}, as is traditional in the
morphological inflection literature, even though the UD treebanks on which we evaluate are
\emph{token-level} resources. Concretely, we compile an incomplete
type-level morphological lexicon from the token-level resource. To create this
resource, we gather all unique form-lemma-tag triples $\langle f, \ell,
m \rangle$ present in the UD test data.\footnote{Some of these form-lemma-tag triples will
  overlap with those seen in the training data.}

\subsection{Baselines}\label{sec:baselines}
As mentioned before, most work on morphological inflection has
considered the task of estimating statistical inflectors from
type-level lexicons. Here, in contrast, we require token-level
annotation to estimate our model. For this reason, there is neither
a competing approach whose numbers we can make a fair comparison to nor is there
an open-source system we could easily run in the token-level setting.
This is why we treat our token-level data as a list of ``types''%
\footnote{Typical type-based inflection lexicons are likely not i.i.d. samples from natural utterances, but we have no other choice if we want to make use of only our token-level data and not additional resources like frequency and regularity of forms.}
and then use two simple type-based baselines.

First, we consider the probabilistic
finite-state transducer used as the baseline for the CoNLL-SIGMORPHON
2017 shared task.\footnote{\url{https://sites.google.com/view/conll-sigmorphon2017/}} We
consider this a relatively strong baseline, as we seek to generalize
from a minimal amount of data. As described by
\newcite{cotterell-conll-sigmorphon2017}, the baseline performed quite
competitively in the task's low-resource setting. Note that the
finite-state machine is created by heuristically extracting prefixes
and suffixes from the word forms, based on an unsupervised alignment
step. The second baseline is our neural inflector $p(f \mid \ell, m)$
given in \cref{sec:parameterization} \emph{without} the
semi-supervision; this model is state-of-the-art on the 
high-resource version of the task.

We will refer to our baselines as follows:
\textbf{FST} is the probabilistic transducer, \textbf{NN} is the
neural sequence-to-sequence model \emph{without} semi-supervision, and
\textbf{SVAE} is the structured variational autoencoder, which is
equivalent to \textbf{NN} but also trained using wake-sleep and
unlabeled data.

\subsection{Results}\label{sec:results}
We ran the three models on 23 languages with the hyperparameters
and experimental details described in \cref{sec:hyperparameters}.
We present our results in \cref{fig:violin} and in \cref{tab:results}. We also provide
sample output of the generative model created using the dream step in \cref{sec:fake}. The high-level take-away
is that on almost all languages we are able to exploit
the unlabeled data to improve the sequence-to-sequence model using
unlabeled data, i.e., SVAE outperforms the NN model on \emph{all} languages
across \emph{all} training scenarios.
However, in many cases, the FST model
is a better choice---the FST can sometimes generalize better from a handful of
supervised examples than the neural network, even with semi-supervision (SVAE).
We highlight three finer-grained observations below.

\paragraph{Observation 1: FST Good in Low-Resource.}
As clearly evinced in \cref{fig:violin}, the baseline FST is still competitive with the NN, or
even our SVAE when data is extremely scarce.
Our neural architecture is quite general, and lacks the prior knowledge and inductive biases of the rule-based system, which become more pertinent in low-resource scenarios.
Even though our semi-supervised strategy clearly improves the performance of NN, we cannot always recommend SVAE for the case when we only have 500 annotated tokens, but on average it does slightly better.
The SVAE surpasses the FST when moving up to 1000 annotated tokens, becoming even more pronounced at 5000 annotated tokens.

\paragraph{Observation 2: Agglutinative Languages.}
The next trend we remark upon is that languages of an agglutinating
nature tend to benefit more from the semi-supervised learning. Why
should this be?  Since in our experimental set-up, every language sees
the \emph{same number of tokens}, it is naturally harder to generalize
on languages that have more distinct morphological variants.
Also, by the nature of agglutinative languages, relevant morphemes could be arbitrarily far from the edges of the string, making the (NN and) SVAE's ability to learn more generic rules even more valuable.

\paragraph{Observation 3: Non-concatenative Morphology.}
One interesting advantage that the neural models have over the FSTs is the ability to learn non-concatenative phenomena. The FST model is based on prefix and suffix rewrite rules and, naturally, struggles when the correctly reinflected form is more than the concatenation of these parts. Thus we see that for the two semitic language, the SVAE is the best method across all resource settings.

\section{Conclusion}
We have presented a novel generative model
for morphological inflection generation in context.
The model allows us to exploit unlabeled data in the training
of morphological inflectors. As the model's rich parameterization
prevents tractable inference, we craft a variational inference
procedure, based on the wake-sleep algorithm, to marginalize out
the latent variables. Experimentally, we provide empirical
validation on 23 languages. We find that, especially in the
lower-resource conditions, our model improves by large margins
over the baselines.

\bibliography{structured-autoencoder}
\bibliographystyle{acl_natbib}

\newpage
\newpage
\newpage
\appendix
\section{Hyperparameters and Experimental Details}\label{sec:hyperparameters}
Here, we list all the hyperparameters and other experimental
details necessary for the reproduction of the numbers presented in
\cref{tab:results}. The final experiments were produced with the
follow setting. We performed a modest grid search over various
configurations in the search of the best option on development
for each component. 

\paragraph{LSTM Morphological Tag Language Model.}
The morphological tag language model is a $2$-layer vanilla LSTM trained with hidden size of $200$.
It is trained to for $40$ epochs using SGD with a cross entropy loss objective, and an initial learning rate of $20$ where the learning rate
is quartered during any epoch where the loss on the validation set reaches a new minimum.
We regularize using dropout of $0.2$ and clip gradients to $0.25$.
The morphological tags are embedded (both for input and output) with a multi-hot encoding into $\mathbb{R}^{200}$,
where any given tag has an embedding that is the sum of the embedding for its constituent POS tag and each of its constituent slots.

\paragraph{Lemmata Generator.}
The lemma generator is a single-layer vanilla LSTM, trained for 10000 epochs using SGD with a learning rate of 4, using a batch size of 20000.
The LSTM has 50 hidden units, embeds the POS tags into $\mathbb{R}^5$ and each token (i.e., character) into $\mathbb{R}^5$.
We regularize using weight decay (1e-6), no dropout, and clip gradients to 1.
When sampling lemmata from the model, we cool the distribution using a temperature of 0.75 to generate more ``conservative'' values.
The hyperparameters were manually tuned on Latin data to produce sensible output and fit development data and then reused for all languages of this paper.

\paragraph{Morphological Inflector.}
The reinflection model is a single-layer GRU-cell seq2seq model with a bidirectional encoder and multiplicative attention in the style of \newcite{luong-pham-manning:2015:EMNLP}, which we train for 250 iterations of AdaDelta~\citep{adadelta}.  Our search over the remaining hyperparameters was as follows (optimal values in bold): input embedding size of $[$50, 100, \textbf{200}, 300 $]$, hidden size of $[$50, \textbf{100}, 150, 200$]$, and a dropout rate of $[$0.0, 0.1, 0.2, 0.3, 0.4, \textbf{0.5}$]$.

\paragraph{Lemmatizer and Morphological Tagger.}
\label{par:lemmatizer_and_morph_tagger}
The joint lemmatizer and tagger is \lemming{} as described in \cref{sub:our_variational_family}.
It is trained with default parameters, the pretrained word vectors from \newcite{bojanowski2016enriching} as type embeddings, and beam size $3$.

\paragraph{Wake-Sleep}
We run two iterations ($I=2$) of wake-sleep. Note that each of the
subparts of wake-sleep: estimating $\ptheta$ and estimating
$\qphi$ are trained to convergence and use the hyperparameters
described in the previous paragraphs.  We set $\gamma_\textit{wake}$
and $\gamma_\textit{sleep}$ to $0.25$, so we observe roughly
$\sfrac{1}{4}$ as many dreamt samples as true samples. The samples
from the generative model often act as a regularizer, helping the
variational approximation (as measured on morphological tagging and
lemmatization accuracy) on the UD development set, but sometimes the
noise lowers performance a mite. Due to a lack of space in the initial paper, we did
not deeply examine the performance of the tagger-lemmatizer outside the context
of improving inflection prediction accuracy. Future work will investigate question
of how much tagging and lemmatization can be improved through the incorporation
of samples from our generative model. In short, our efforts will evaluate the inference
network in its own right, rather than just as a variational approximation to the posterior.

\section{Fake Data from the Sleep Phase}\label{sec:fake}
An example sentence $\tilvf$ sampled via $\angles{\tilvf, \tilvell, \tilvm} \sim \ptheta\left(\cdot, \cdot, \cdot\right)$ in Portuguese:

\noindent
\texttt{dentremeticamente » isso Procusas Da Fase » pos a acord\'itica M\'aisringeringe Ditudis A ana , Urevirao Da De O linsith.muital , E que chegou interalionalmente Da anundica De m\^epinsuriormentais .}

\noindent
and in Latin:

\noindent
\texttt{inpremcret ita sacrum super annum pronditi avocere quo det tuam nunsidebus quod puela ?}

\end{document}